\documentclass{article} %
\usepackage{iclr2025_conference,times}

\usepackage{amsmath,amsfonts,bm}

\def\eqref#1{equation~\ref{#1}}

\def\1{\bm{1}}

\def\ve{{\bm{e}}}

\def\vk{{\bm{k}}}

\def\vq{{\bm{q}}}

\def\vv{{\bm{v}}}

\def\vx{{\bm{x}}}
\def\vy{{\bm{y}}}

\def\mA{{\bm{A}}}
\def\mB{{\bm{B}}}

\def\mD{{\bm{D}}}

\def\mH{{\bm{H}}}
\def\mI{{\bm{I}}}

\def\mP{{\bm{P}}}

\def\mX{{\bm{X}}}

\DeclareMathAlphabet{\mathsfit}{\encodingdefault}{\sfdefault}{m}{sl}
\SetMathAlphabet{\mathsfit}{bold}{\encodingdefault}{\sfdefault}{bx}{n}

\newcommand{\R}{\mathbb{R}}

\usepackage{hyperref}
\usepackage{url}
\usepackage{listings}
\usepackage{xcolor}
\usepackage{adjustbox}
\usepackage{cleveref}
\usepackage{tikz}
\usepackage{pgfplots}
\usepackage{amssymb}
\usepackage{booktabs}
\usepackage{makecell}
\usepackage{tcolorbox}
\usetikzlibrary{automata,angles,quotes,calc, matrix, positioning,decorations.pathreplacing,shadows,shapes, arrows.meta}

\usepackage{enumitem}

\usepackage{subcaption} %
\usepackage{graphicx}   %
\usetikzlibrary{shapes.geometric} %

\definecolor{ipythonprompt}{RGB}{40,40,40}
\definecolor{ipythonout}{RGB}{200,0,0}
\definecolor{pythonstring}{RGB}{186,85,211}
\definecolor{pythonkeyword}{RGB}{0,0,255}
\definecolor{pythoncomment}{RGB}{128,128,128}

\definecolor{modernblue}{RGB}{52, 152, 219}      %
\definecolor{moderngreen}{RGB}{46, 204, 113}     %
\definecolor{modernorange}{RGB}{230, 126, 34}    %
\definecolor{modernpurple}{RGB}{155, 89, 182}    %
\definecolor{modernred}{RGB}{231, 76, 60}        %
\definecolor{modernyellow}{RGB}{241, 196, 15}    %
\definecolor{moderngray}{RGB}{149, 165, 166}     %
\definecolor{modernlightgray}{RGB}{236, 240, 241} %
\definecolor{moderndarkgray}{RGB}{52, 73, 94}    %
\definecolor{opTrans}{RGB}{70, 130, 180}   %
\definecolor{opQuery}{RGB}{255, 140, 0}    %
\definecolor{opReveal}{RGB}{34, 139, 34}   %
\definecolor{failRed}{RGB}{178, 34, 34}    %
\definecolor{safeGreen}{RGB}{60, 179, 113} %

\definecolor{ghostRed}{RGB}{255, 235, 235} %
\definecolor{textRed}{RGB}{178, 34, 34}    %
\definecolor{revealGreen}{RGB}{235, 255, 235} %
\definecolor{textGreen}{RGB}{34, 139, 34}     %

\lstdefinestyle{ipython}{
    language=Python,
    basicstyle=\ttfamily\footnotesize,
    keywordstyle=\color{pythonkeyword}\bfseries,
    stringstyle=\color{pythonstring},
    commentstyle=\color{pythoncomment}\itshape,
    showstringspaces=false,
    breaklines=true,
    frame=single,
    rulecolor=\color{black!30},
    backgroundcolor=\color{white!5},
    xleftmargin=0.2cm,
    xrightmargin=0.2cm,
    aboveskip=1em,
    belowskip=1em,
    linewidth=0.275\linewidth, %
    moredelim=[s][\color{ipythonprompt}\bfseries]{In [}{]:},
    moredelim=[s][\color{ipythonout}]{Out[}{]:},
    literate={'}{\textquotesingle}1
}

\usepackage[symbol]{footmisc} %
\usepackage{wrapfig}

\newtheorem{definition}{Definition}

\makeatletter
\renewcommand\@fnsymbol[1]{%
  \ifcase#1\or\dagger\or\ddagger\or\mathsection\or\mathparagraph\or\|\or\#\fi}
\makeatother
\newcommand{\norm}[1]{\left\lVert#1\right\rVert}

\usepackage[textwidth=1.3in,textsize=tiny]{todonotes}

\newtcolorbox{takeawaybox}[1]{
  colback=modernlightgray!30,
  colframe=moderndarkgray,
  boxrule=0.5pt,
  arc=2pt,
  left=6pt,right=6pt,top=4pt,bottom=4pt,
  fonttitle=\bfseries,
  title={#1}
}

\title{Learning State-Tracking from Code\\ Using Linear RNNs}

\author{Julien Siems$^{*\heartsuit\diamondsuit\dagger}$,~~Riccardo Grazzi$^{*\clubsuit}$,~~Korbinian Pöppel$^{\bigstar}$,~~Kirill Kalinin$^{\clubsuit}$, \\ \textbf{Hitesh Ballani$^{\clubsuit}$,~~Babak Rahmani$^{*\clubsuit\spadesuit}$} \\
Equal contribution$^*$, University of Freiburg$^{\heartsuit}$, Microsoft Research$^{\clubsuit}$, Prior Labs$^{\diamondsuit}$,\\ ELLIS Institute Tübingen$^{\bigstar}$, T\"ubingen AI Center$^\spadesuit$\\
{\small $^{\dagger}$Work done during an internship at Microsoft Research.} \\
{\small \texttt{juliensiems@gmail.com}}
\quad
{\small \texttt{t-rgrazzi@microsoft.com}}
\quad
{\small \texttt{rahmani.b91@gmail.com}}
}

\iclrfinalcopy %
\begin{document}

\maketitle
\begin{abstract}
Over the last years, state-tracking tasks, particularly permutation composition, have become a testbed to understand the limits of sequence-to-sequence architectures like Transformers and RNNs (linear and non-linear). However, these are often sequence-to-sequence tasks: learning to map actions (permutations) to states, which is incompatible with the next-token prediction setting commonly used to train language models. We address this gap by converting permutation composition into code via REPL traces that interleave state-reveals through prints and variable transformations. We show that linear RNNs capable of state-tracking excel also in this setting, while Transformers still fail. Motivated by this representation, we investigate why tracking states in code is generally difficult: actions are not always fully observable. We frame this as tracking the state of a probabilistic finite-state automaton with deterministic state reveals and show that linear RNNs can be worse than non-linear RNNs at tracking states in this setup. 
\end{abstract}
\section{Introduction}\label{sec:introduction}
\vspace{-3mm}
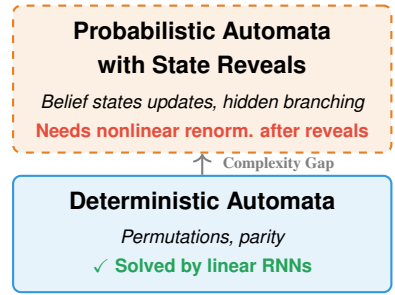
\begin{wrapfigure}[18]{r}{0.4\textwidth}
\vspace{-4.3mm}
\centering
\begin{tikzpicture}[
    font=\sffamily,
    every node/.style={align=center},
    box/.style={draw, thick, rounded corners=3pt, minimum width=0.36\textwidth, inner sep=6pt},
]
\node[box, fill=modernorange!12, draw=modernorange, dashed] (prob) {
    {\bfseries\small Probabilistic Automata }\\[2pt]
    {\bfseries\small with State Reveals }\\[2pt]
    {\scriptsize\itshape Belief states updates, hidden branching}\\
    {\scriptsize\bfseries\color{modernred}  Needs nonlinear renorm. after reveals}
};
\node[box, fill=modernblue!12, draw=modernblue, below=8pt of prob] (det) {
    {\bfseries\small Deterministic Automata}\\[2pt]
    {\scriptsize\itshape Permutations, parity}\\    {\scriptsize\bfseries\color{moderngreen!80!black} \checkmark\ Solved by linear RNNs}
};
\draw[->, thick, gray] (det.north) -- (prob.south)
    node[midway, right=4pt, font=\tiny\bfseries, text=gray] {Complexity Gap};
\end{tikzpicture}
\caption{\textbf{The State-Tracking Hierarchy.} Linear RNNs solve deterministic tasks (blue) but struggle with probabilistic automata (orange). Real-world code often requires belief-state tracking in the orange region.}
\label{fig:tracking_hierarchy}
\vspace{-4mm}
\end{wrapfigure}
State-tracking is fundamental across many domains: In order to understand a program state, models must track variable states during code execution~\citep{team2025cwm}, board configurations in game-playing~\citep{toshniwal2022chess, harang2025tracking}, and environment representations in world-modeling~\citep{vafa2024evaluating, vafa2025world}. Theoretical work has established a divide between associative recall, where transformers excel, and state-tracking, where recurrent neural networks perform well~\citep{merrill2020formal, merrill2023parallelism}. Recently, linear RNNs like Mamba~\citep{gumamba,dao2024transformers} and DeltaNet~\citep{yang2024parallelizing,yanggateddeltanet} were introduced which allow parallelization across the sequence length. While early linear RNNs were incapable of complex state-tracking~\citep{merrill2024illusion, sarrof2024expressive}, extending the eigenvalues of the state-transition matrix from $[0, 1]$ to $[-1, 1]$~\citep{grazziunlocking,siems2025deltaproduct, peng2025rwkv}, or introducing recurrent fixed-point self-iteration over linear RNNs enables solving permutation composition ~\citep{schoneimplicit}. Yet empirical gains on real-world tasks remain modest~\citep{jellyfish,grazziunlocking}. Concurrently, work on parallelizing nonlinear RNNs has emerged~\citep{limparallelizing, gonzalez2024towards, danieli2025pararnn}, but benchmarking these architectures solely on tasks expressible as Deterministic Finite-State Automata (DFA) obscures their potential, since linear RNNs already solve such tasks efficiently~\citep{peng2025rwkv}.

To understand how state-tracking can be learned by language models, we study it through the lens of next-token prediction. Current benchmarks typically use a sequence-to-sequence setup—mapping sequences of actions to states, which deviates significantly from the objective used to train language models. In this work, we address this gap and present the following contributions:

\begin{itemize}[leftmargin=*]
    \item \textit{State-Tracking via Next-Token Prediction:} We convert permutation composition into code via Python REPL traces that interleave variable transformations with partial state-reveals (print statements). This creates a realistic setting where state-tracking must be learned via next-token prediction rather than sequence-to-sequence supervision.
    
    \item \textit{Linear RNNs vs. Transformers:} We show that linear RNNs capable of state-tracking (specifically DeltaNet with extended eigenvalues) excel in this setting even with sparse supervision, whereas Transformers fail to generalize.
    
    \item \textit{Probabilistic State-Tracking Limits of linear RNNs:} We investigate why tracking states in real code is strictly harder than permutations: actions are often not fully observable. We frame this as tracking the state of a \textit{Probabilistic Finite-State Automaton with State Reveals} (PFSA-SR) and show concrete adversarial sequences under which the natural linear RNN representations of belief states suffer exponential norm decay.
\end{itemize}

\section{Related Work}\label{sec:related_work}
\paragraph{Deterministic state tracking.}
Prior work has largely studied state tracking in \emph{deterministic} settings, regular languages and finite-state automaton (FSA) emulation, using length generalization as the primary stress test \citep{hahn2020theoretical,bhattamishra2020ability,deletang2022neural,liu2022transformers}. These studies reveal sharp, architecture-dependent generalization gaps across the Chomsky hierarchy \citep{deletang2022neural} and characterize when sequence models can implement finite-state computation under bounded depth and precision \citep{merrill2024illusion,terzic2025structured}. However, architectural improvements on synthetic benchmarks translate to only modest gains in language-model evaluations \citep{grazziunlocking,siems2025deltaproduct,jellyfish}. Moreover, widely used setups such as group-word problems \citep{liu2022transformers} differ substantially from code execution, where supervision is sparse and state evolution often stochastic. Moreover, the standard group-word problem benchmarks are sequence-to-sequence: inputs are sequence of actions and outputs are sequences of states. This differs significantly from the next-token prediction paradigm used to train modern language models. 
We address this gap by studying probabilistic state tracking in code-like data, in particular REPL traces, which interleave actions with partial state reveals in the same sequence.
Recent work in reinforcement learning under partial observability~\citep{luisuncertainty, prasanna2025one} and language modeling~\citep{shaj2026kalman} empirically confirms that standard linear recurrent models struggle when uncertainty reasoning is required, motivating explicit probabilistic mechanisms such as Kalman filtering. Our PFSA-SR framework provides a theoretical explanation for these empirical observations.

\paragraph{Transformers and linear RNN expressivity.}
Transformer expressivity analyses failure modes on state-tracking tasks including shortcut learning and poor length extrapolation \citep{hahn2020theoretical,bhattamishra2020ability,merrill2024illusion,deletang2022neural,liu2023code,strobl2024formal}. In parallel, work on efficient recurrent and SSM-style architectures has explored structured transition parameterizations trading efficiency for expressive power \citep{schlag2021learning,yang2024parallelizing,fan2024advancing,walker2025structured,peng2025rwkv}, showing that richer spectra or factorizations unlock new state-tracking behaviors without changing asymptotic cost \citep{grazziunlocking,siems2025deltaproduct}. These works primarily address deterministic transitions, abstracting away belief evolution under uncertainty or sparse supervision.

\paragraph{Probabilistic automata and partial observability.}
Classical probabilistic finite-state models—weighted automata, IO-HMMs, POMDPs—formalize uncertainty via belief-state updates under partial observations \citep{RABIN1963230,bengio1994input,aastrom1965optimal,kaelbling1998planning}. Recent work characterizes recurrent models as restricted probabilistic finite-state systems \citep{svete2023recurrent,butoitraining,borenstein2024languages}, but focuses on representational capacity of nonlinear RNNs. Our PFSA-SR instead analyzes the failure mode of \emph{linear} RNNs under partial observability induced by hard, support-pruning reveals (e.g., asserts, prints) that deterministically eliminate inconsistent trajectories, enabling direct comparison of linear and nonlinear dynamics for belief-state updates. Also note the concurrent work by~\citet{zhao2026linear} investigating nearly deterministic state-tracking.

\section{Background}
\vspace{-2mm}
\paragraph{Linear RNNs.} Linear RNNs process input sequences through stacked layers. Each layer transforms input vectors $\vx_1, \dots, \vx_t\in\R^l$ into outputs $\hat{\vy}_1,\dots,\hat{\vy}_t\in\R^p$ via the linear recurrence
\begin{equation}\label{eq:linrnn}
\begin{aligned}
\mH_i = \mA(\vx_i) \mH_{i-1} + \mB(\vx_i),\qquad
\hat{\vy}_i = \mathrm{dec}(\mH_i, \vx_i)\qquad
\text{for } i \in \{1, \dots, t\}
\end{aligned}
\end{equation}
where $\mH_0 \in \R^{n \times d}$ is the initial state, $\mA: \R^l \to \R^{n \times n}$ produces the state-transition matrix, $\mB: \R^l \to \R^{n \times d}$ injects new information, and $\mathrm{dec}: \R^{n \times d} \times \R^l \to \R^p$ generates outputs. These functions are learned, with $\mathrm{dec}$ typically containing a feedforward network. Different linear RNN variants differ in their implementations of $\mA$, $\mB$, and $\mathrm{dec}$.

We focus on \textit{DeltaNet}~\citep{schlag2021linear,schlag2021learning}, recently shown to be  parallelizable across the sequence length~\citep{yang2024parallelizing,yanggateddeltanet}. DeltaNet parameterizes the recurrence as
$\mA(\vx_i) = \mI -\beta_i \vk_i\vk_i^\top,\quad\mB(\vx_i)=\beta_i \vk_i\vv_i^\top, \quad\mathrm{dec}(\mH_i, \vx_i)=\psi(\mH_i^\top\vq_i)$
where $\beta_i \in [0,1]$ and $\vq_i, \vk_i \in \R^n$ (with $\norm{\vq_i} =\norm{\vk_i} = 1$), $\vv_i \in \R^d$ are learned functions of $\vx_i$. Here $\mA(\vx_i)$ is a Householder transformation~\citep{householder1958unitary} with eigenvalues 1 (multiplicity $n-1$) and $1-\beta_i$ (multiplicity 1). Restricting eigenvalues to $[0,1]$ limits state-tracking capabilities, but extending to $[-1,1]$ enables complex tracking behaviors like parity and permutation composition~\citep{grazziunlocking}. The linear recurrence enables parallelization across sequence length \citep{blelloch1990prefix, martin2017parallelizing, hua2022transformer,sun2023retentive,yanggated}.

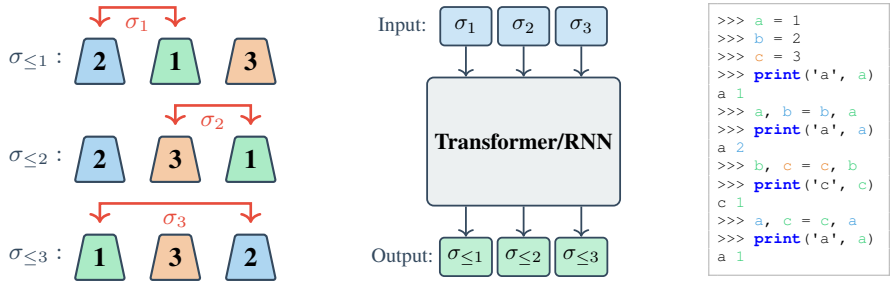
\begin{figure}
    \centering
\vspace{-3pt}%
\hfill
\adjustbox{width=0.27\linewidth, valign=t}{
    \begin{tikzpicture}[scale=1.0]
    \tikzset{
        cup/.style={
            trapezium,
            trapezium left angle=75,
            trapezium right angle=75,
            trapezium stretches=true,
            minimum height=0.6cm,
            minimum width=0.6cm,
            draw=moderndarkgray,
            thick,
            rounded corners=1pt,
            rotate=0,
            path picture={
                \fill[#1!40] ([xshift=-0.2cm, yshift=-0.15cm]path picture bounding box.south west) -- 
                              ([xshift=0.2cm, yshift=-0.15cm]path picture bounding box.south east) -- 
                              ([xshift=0.07cm, yshift=0.15cm]path picture bounding box.north east) --
                              ([xshift=-0.07cm, yshift=0.15cm]path picture bounding box.north west) -- cycle;
            }
        },
        swap arrow/.style={<->, thick, modernred, line width=1pt, shorten >=4pt, shorten <=4pt},
        label/.style={font=\small, color=moderndarkgray}
    }
    
    \node[label] at (-0.85, 0) {$\sigma_{\leq 1}:$};
    \node[cup=modernblue] (cup1-1) at (0,0) {\bfseries 2};
    \node[cup=moderngreen] (cup1-2) at (1,0) {\bfseries 1};
    \node[cup=modernorange] (cup1-3) at (2,0) {\bfseries 3};
    
    \node[label] at (-0.85, -1.3) {$\sigma_{\leq 2}:$};
    \node[cup=modernblue] (cup2-1) at (0,-1.3) {\bfseries 2};
    \node[cup=modernorange] (cup2-2) at (1,-1.3) {\bfseries 3};
    \node[cup=moderngreen] (cup2-3) at (2,-1.3) {\bfseries 1};
    
    \node[label] at (-0.85, -2.6) {$\sigma_{\leq 3}:$};
    \node[cup=moderngreen] (cup3-1) at (0,-2.6) {\bfseries 1};
    \node[cup=modernorange] (cup3-2) at (1, -2.6) {\bfseries 3};
    \node[cup=modernblue] (cup3-3) at (2,-2.6) {\bfseries 2};
    
    \draw[swap arrow] (cup1-1.north) -- ++(0,0.4) -- ++(1,0) node[midway, below] {\small $\sigma_1$} -- (cup1-2.north);
    \draw[swap arrow] (cup2-2.north) -- ++(0,0.4) -- ++(1,0) node[midway, below] {\small $\sigma_2$} -- (cup2-3.north);
    \draw[swap arrow] (cup3-1.north) -- ++(0,0.4) -- ++(2,0) node[midway, below] {\small $\sigma_3$} -- (cup3-3.north);    
\end{tikzpicture}}
\hfill
\adjustbox{width=0.25\linewidth, valign=t}{
\begin{tikzpicture}[
    scale=0.9,
    token/.style={rectangle, draw=moderndarkgray, thick, minimum height=0.6cm, minimum width=0.8cm, rounded corners=2pt},
    model/.style={rectangle, draw=moderndarkgray, thick, minimum height=2cm, minimum width=2.5cm, rounded corners=3pt, fill=modernlightgray},
    arrow/.style={->, thick, moderndarkgray, line width=0.8pt},
    label/.style={font=\small, color=moderndarkgray}
]

\node[label] at (-1.1, 2.5) {Input:};
\node[token, fill=modernblue!25] (in1) at (0, 2.5) {$\sigma_1$};
\node[token, fill=modernblue!25] (in2) at (1, 2.5) {$\sigma_2$};
\node[token, fill=modernblue!25] (in3) at (2, 2.5) {$\sigma_3$};

\node[model] (model) at (1, 0.5) {\textbf{Transformer/RNN}};

\draw[arrow] (in1) -- (in1 |- model.north);
\draw[arrow] (in2) -- (in2 |- model.north);
\draw[arrow] (in3) -- (in3 |- model.north);

\node[label] at (-1.1, -1.5) {Output:};
\node[token, fill=moderngreen!30] (out1) at (0, -1.5) {$\sigma_{\leq 1}$};
\node[token, fill=moderngreen!30] (out2) at (1, -1.5) {$\sigma_{\leq 2}$};
\node[token, fill=moderngreen!30] (out3) at (2, -1.5) {$\sigma_{\leq 3}$};

\draw[arrow] (out1 |- model.south) -- (out1);
\draw[arrow] (out2 |- model.south) -- (out2);
\draw[arrow] (out3 |- model.south) -- (out3);

\end{tikzpicture}}
\hfill
\adjustbox{width=0.17\textwidth, valign=t}{\begin{minipage}[t]{0.68\linewidth}
\vspace{-3mm}
\begin{lstlisting}[style=ipython, basicstyle=\ttfamily\tiny]
>>> @\textcolor{moderngreen}{a}@ = 1
>>> @\textcolor{modernblue}{b}@ = 2
>>> @\textcolor{modernorange}{c}@ = 3
>>> print('a', @\textcolor{moderngreen}{a}@)
a @\textcolor{moderngreen}{1}@
>>> @\textcolor{moderngreen}{a}@, @\textcolor{modernblue}{b}@ = @\textcolor{modernblue}{b}@, @\textcolor{moderngreen}{a}@
>>> print('a', @\textcolor{modernblue}{a}@)
a @\textcolor{modernblue}{2}@
>>> @\textcolor{moderngreen}{b}@, @\textcolor{modernorange}{c}@ = @\textcolor{modernorange}{c}@, @\textcolor{moderngreen}{b}@
>>> print('c', @\textcolor{moderngreen}{c}@)
c @\textcolor{moderngreen}{1}@
>>> @\textcolor{modernblue}{a}@, @\textcolor{moderngreen}{c}@ = @\textcolor{moderngreen}{c}@, @\textcolor{modernblue}{a}@
>>> print('a', @\textcolor{moderngreen}{a}@)
a @\textcolor{moderngreen}{1}@
\end{lstlisting}\end{minipage}}
\hspace{10mm}
\vspace{-3mm}

\caption{Three representations of the permutation tracking task. \textbf{Left:} The shell game analogy showing cups being swapped to track object positions. \textbf{Center:} The sequence-to-sequence modeling approach from \citet{merrill2024illusion}, where a Transformer/RNN processes input permutations $\sigma_i$ and outputs cumulative states $\sigma_{\leq i} = \prod_{j=1}^i \sigma_j$ at each position. \textbf{Right:} Our code-based representation using Python REPL traces, where variable swaps implement permutations and print statements reveal partial cumulative states for next-token prediction training.}\label{fig:repl_example}
\end{figure}

\section{From Permutation Groups to Variable Tracking in Code}\label{sec:motivation}
\vspace{-2mm}
We translate the sequence-to-sequence setup for group-word problems from \citet{merrill2024illusion} to next-token prediction. This setup has been influential in understanding the state-tracking abilities of recurrent model \citep{schoneimplicit,grazziunlocking,siems2025deltaproduct,movahedifixed} and parallels the shell game where cups containing objects are shuffled.

\begin{wrapfigure}[13]{r}{0.35\textwidth} 
    \centering
    \vspace{-17pt} 
    \includegraphics[width=1.0\linewidth]{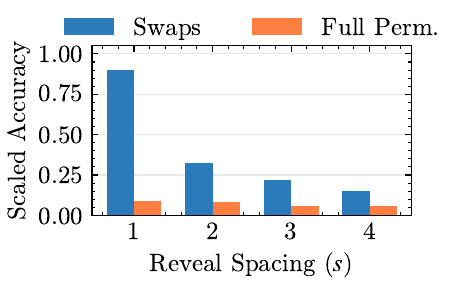}
    \vspace{-6mm}
    \caption{Transformers require dense state supervision to solve REPL traces. Accuracy drops as reveal spacing increases, with full permutations failing under any sparsity.}  \label{fig:transformer_architecture}
    \vspace{-5pt} %
\end{wrapfigure}
\textbf{Sequence-to-sequence modeling:} For permutation group $S_n$, we sample input permutations $\sigma_{\textnormal{IN}} = [\sigma_1, \ldots, \sigma_{m}]$. At position $i\in [1,\dots,m]$, the model predicts the cumulative state $\sigma_{\leq i}=\prod_{j=1}^i\sigma_j$ with labels $\sigma_{\textnormal{OUT}}=[\sigma_{\leq 1}, \dots, \sigma_{\leq m}]$. This provides dense supervision by computing loss at every position.

\textbf{Next-token prediction (NTP):} We adapt permutation groups for NTP using Python REPL traces~\citep{DeutschBerkeley1964,van1995python} that demonstrate variable shuffling. \Cref{fig:repl_example} shows an example for $S_3$. We interleave commands with print statements revealing partial states rather than only end of sequence states, providing denser signal while remaining realistic (similar to logging during execution). Full reveals would trivialize the task by allowing the model to ignore prior permutations.

\subsection{Experiments}
\textbf{Transformers require dense state supervision for state-tracking.}
To examine how supervision density affects state-tracking in pretrained Transformer LLMs, we finetune a Qwen3-0.6B-Base~\citep{qwen3technicalreport} model using standard NTP on Python REPL traces of 64 commands and $S_5$. We consider both elementary swaps and full permutations, and vary the reveal spacing from 1 to 4, thereby progressively reducing the density of explicit state information available during training. As shown in~\Cref{fig:transformer_architecture}, Transformers rely critically on dense supervision: as reveal spacing increases and state information becomes sparser, the model rapidly loses the ability to track permutations. For full permutations the model does not learn to solve the task.

\begin{figure}
    \centering
    \adjustbox{width=0.9\textwidth}{
    \includegraphics[width=0.38\linewidth]{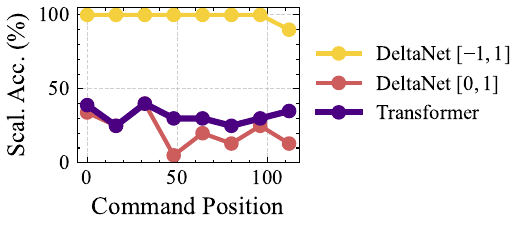}
    \hspace{5mm}
    \includegraphics[width=0.45\linewidth]{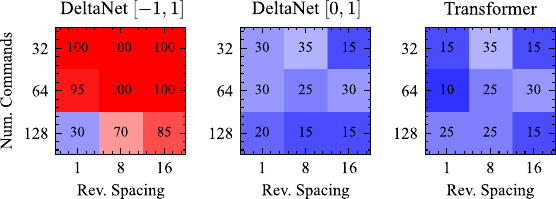}}
    \vspace{-2mm}
    \caption{\textbf{Left:} Per reveal position accuracy averaged across 5 seeds, reveal spacing 16 and num commands 128. Only DeltaNet $[-1, 1]$ manages to reliably learn to perform state tracking. \textbf{Right:} Final reveal position accuracy when increasing the reveal spacing and num. commands beyond training regime, 8 and 64 respectively.}
    \label{fig:architecture_comparison}
\end{figure}

\begin{figure}
    \centering
    \adjustbox{width=1.0\textwidth}{
    \centering
    \includegraphics[width=0.3\linewidth]{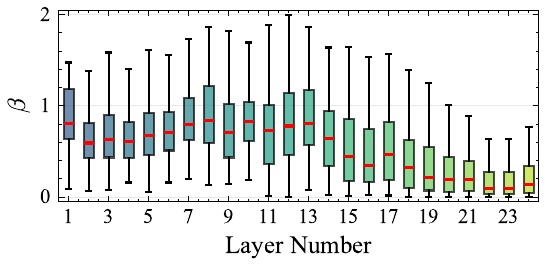}
    \includegraphics[width=0.217\linewidth]{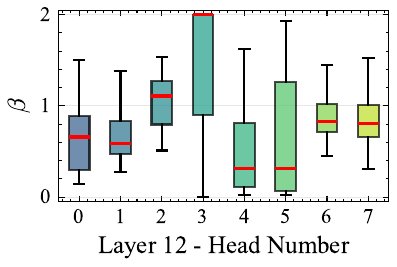}
    \includegraphics[width=0.223\linewidth]{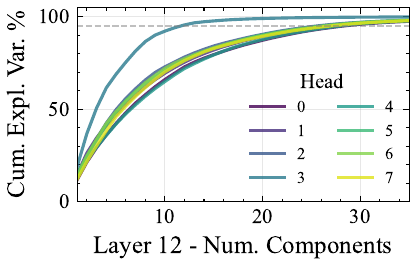}}
    \vspace{-7mm}
    \caption{Interpretability Analysis: Distribution of $\beta$ for a sequence of 512 commands with reveal spacing 4. \textbf{Left:} $\beta$ values aggregated per layer across heads. \textbf{Middle:} $\beta$ per head for layer 12; head 3 stands out as the only head in the network to consistently set its $\beta$ values to 2, making it the state-tracking head. \textbf{Right:} Cumulative explained variance of the PCA of the keys. Head 3 again stands out, being explainable with much fewer components.}
    \label{fig:beta_distribution_across_layers}
\end{figure}

\begin{wrapfigure}[15]{r}{0.35\columnwidth}
\vspace{-14pt}
\centering
\includegraphics[width=1.0\linewidth, trim=0 0 0 0, clip]{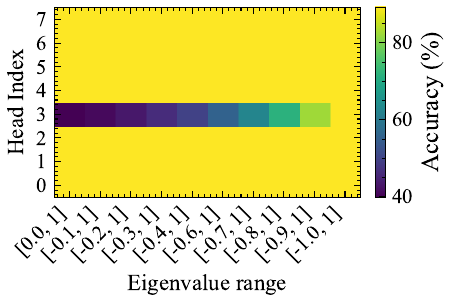}
\vspace{-15pt}
\caption{Intervention analysis: Scaling the $\beta$ range per head in layer 12. State prediction accuracy degrades only when head 3 is scaled, confirming it as the state-tracking head.}
\label{fig:interpretability}
\vspace{-50pt}
\end{wrapfigure}

\textbf{Architecture Determines Whether State-Tracking Extrapolates.}
\label{subsec:motivation_arch}
We train DeltaNet $[-1,1]$, a state-tracking capable architecture, and two non-state-tracking capable architectures DeltaNet $[0,1]$, and Transformer models (all with 280M parameters) on a curriculum of 15{,}000 REPL traces from $S_5$, with 8, 16, 32, and 64 commands and reveal spacings of 1, 2, 4, and 8. We use full permutations rather than elementary swaps, to get a clearer signal on the difference between transformer models and linear RNNs.

The length extrapolation results in~\Cref{fig:architecture_comparison} (averaged across five seeds) show a clear architectural separation: DeltaNet $[-1,1]$ learns to state-track perfectly and extrapolates reliably across all seeds, whereas the Transformer does not learn to perform state-tracking even when trained from scratch on these sequences. 

\textbf{Interpretability Analysis.} \citet{grazziunlocking} showed that to learn state-tracking, at least one head must learn to set $\beta$ values to 2 in one of the layers (to obtain an eigenvalue of -1) and find appropriate keys in the right subspace. This was previously demonstrated for a single layer DeltaProduct$_2$ model by~\citet{siems2025deltaproduct} for the $S_4$ group. 
We verify whether DeltaNet$[-1,1]$ learns to perform state-tracking using NTP by leveraging its extended eigenvalue range. Therefore, we pass sequences of 512 commands with spacing 4 through the model and retrieve the $\beta$ values using NNsight~\citep{FiottoKaufman2024NNsightAN}. Figure~\ref{fig:beta_distribution_across_layers} shows the distribution of $\beta$ across layers and heads. We observe that many $\beta$ values exceed 1, which results in negative eigenvalues for the generalized Householder. Layer 12 stands out, with at least one head consistently estimating most $\beta$ values at 2 (Fig.~\ref{fig:beta_distribution_across_layers}, middle). Furthermore, PCA analysis reveals that the keys of this head lie in a significantly lower-dimensional linear subspace than the keys of any other head in the network. We hypothesize that this head has learned to be the state-tracking head of the network. Results for the remaining layers are  in~\Cref{app:interpretability}.

To confirm that head 3 in layer 12 is the state-tracking head, we perform an intervention where we gradually scale $\beta$ to restrict the eigenvalue range from $[-1, 1]$ to $[0, 1]$ for each head independently with results shown in Fig.~\ref{fig:interpretability}. We evaluate on sequences with 5 variables, reveal spacing 2, and 64 commands. Scaling any head except head 3 leaves the state prediction accuracy unchanged. However, scaling head 3 causes significant performance degradation, confirming that it alone enables state-tracking for the model.
\section{Why State-Tracking Fails on Real Code: From Deterministic to Probabilistic Automata with State Reveals}
\label{sec:real_code_failure}

The preceding experiments establish a clear recipe for learning state-tracking via next-token prediction: reveal intermediate states through print statements (for learnability) and use an architecture with state-tracking capabilities. DeltaNet$[-1,1]$ satisfies both criteria and generalizes reliably on synthetic permutation traces (\S\ref{subsec:motivation_arch}). However, this success relies on an idealization that real code violates: \emph{every transition is fully observable}. In our REPL traces, the model sees exactly which variables are swapped at each step. Real code execution rarely gives such transparency.

\begin{figure}[t]
    \centering
    \includegraphics[width=\textwidth, trim=80 150 80 20, clip]{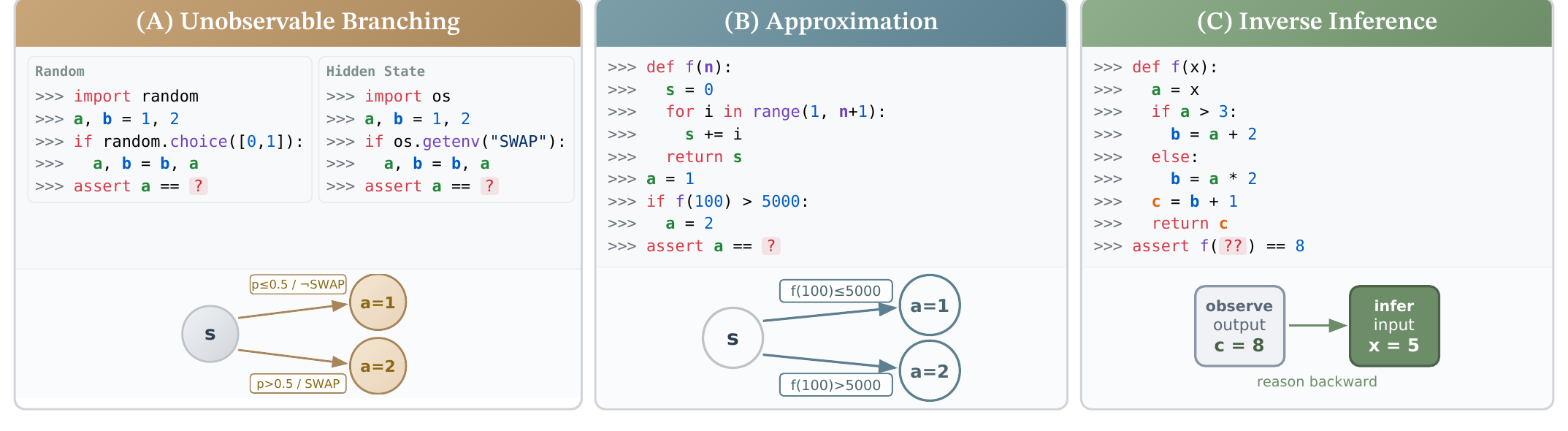}
    \vspace{-7mm}
\caption{Sources of probabilistic state transitions in tracking variables during code execution.
    (A) Unobservable branching from randomness or hidden state.
    (B) Approximation when skipping expensive computations.
    (C) Inverse inference when recovering inputs from outputs.
}
    \label{fig:probabilistic-state-transitions}
\end{figure}

Figure~\ref{fig:probabilistic-state-transitions} illustrates three representative sources of \emph{transition uncertainty} that arise when language models execute code. \textit{(A) Unobservable branching} occurs when code branches on conditions inaccessible to the model, whether from explicit randomness, hidden environment state, or external API calls. \textit{(B) Approximation} arises when models skip expensive operations like loops or recursion, trading exact computation for efficiency but introducing uncertainty about intermediate states. \textit{(C) Inverse inference} tasks, such as CRUXEval-Input~\citep{gu2024cruxeval} or symbolic execution~\citep{king1976symbolic}, require recovering inputs from outputs, which often admits multiple valid solutions.

In all three cases, a model cannot deterministically compute the next state but must instead maintain a \emph{distribution} over possible states. This probabilistic setting differs fundamentally from the deterministic permutation tracking studied in prior work on linear RNNs~\citep{sarrof2024expressive, merrill2024illusion, grazziunlocking, terzic2025structured}. While probabilistic state-tracking has been analyzed for nonlinear RNNs~\citep{svete2023recurrent, butoitraining, borenstein2024languages}, its implications for linear RNNs remain unexplored.

\subsection{Probabilistic Finite-State Automata with State Reveals}

To analyze this formally, we introduce a model that explicitly represents probabilistic transitions (e.g.\@ conditional array swaps) coupled with partial state reveals (e.g.\@ outputs of print statements of the array elements). This formulation treats the input sequence as a series of operations where partial state observation precedes state evolution.

\begin{definition}[PFSA-SR]
A \emph{Probabilistic Finite-State Automaton with State Reveals} (PFSA-SR) is a tuple $(\mathcal{Q}, \Sigma, \delta, \rho, q_0)$ where $\mathcal{Q}$ is a finite set of states with initial state $q_0 \in \mathcal{Q}$, $\Sigma$ is an input alphabet, $\delta: \mathcal{Q} \times \Sigma \to \mathrm{Dist}(\mathcal{Q})$ is the probabilistic transition kernel, and $\rho: \Sigma \to 2^{\mathcal{Q}}$ is the reveal function mapping inputs to subsets of $\mathcal{Q}$.
The system evolves sequentially: at each time-step $t$, the environment (which has access to the true state $q_t$) selects an input symbol $\sigma_t \in \Sigma$ subject to the \emph{consistency constraint} $q_t \in \rho(\sigma_t)$. The next state is then sampled from the transition kernel, $q_{t+1} \sim \delta(q_t, \sigma_t)$.
\end{definition}

The consistency constraint is the defining feature of the reveal mechanism: since the environment must choose $\sigma_t$ so that $q_t \in \rho(\sigma_t)$, the observer (who sees only $\sigma_t$) can eliminate all states outside $\rho(\sigma_t)$ from the current belief. This models code-level observations such as \texttt{print} statements and \texttt{asserts}, which constrain but do not fully determine the latent state.
Two special cases are worth noting: (1) \emph{reveal-only symbols}, where $\delta(q,\sigma)$ places all mass on $q$ for every $q$ (i.e.\@ the state does not change), and (2) \emph{transition-only symbols}, where $\rho(\sigma) = \mathcal{Q}$ (i.e.\@ the reveal is vacuous).

\textbf{Belief Update.} If we only observe the sequence of symbols, the uncertainty in the transitions prevents us from directly tracking the state. What we can model is instead the conditional probability distribution of the state given the input history, denoted by $p(q_t \mid \sigma_1, \dots, \sigma_t)$, which we refer to as the belief $b_t$. If we know $\delta$ and $\rho$, we can update the belief in two stages. First, the reveal step conditions the current belief $b_t$ on the reveal $\rho(\sigma_t)$, since we are guaranteed that $q_t \in \rho(\sigma_t)$. The intermediate belief $b'_t$ is computed by zeroing out states inconsistent with $\rho(\sigma_t)$ and renormalizing. Second, the transition step computes the next belief $b_{t+1}$ by propagating the corrected belief $b'_t$ through the transition kernel:
\[
b'_{t}(q) = b_{t}(q) \cdot \mathbb{I}[q \in \rho(\sigma_t)] / \textstyle\sum_{k \in \rho(\sigma_t)} b_{t}(k) \qquad b_{t+1}(q') = \textstyle\sum_{q \in \mathcal{Q}} b'_t(q) \cdot \delta(q, \sigma_t)(q')
\]
Let $m = |\mathcal{Q}|$ be the number of states. We represent the belief $b_t$ as a column vector in the unit simplex $\Delta_m \subset \mathbb{R}^m$. For each input $\sigma \in \Sigma$, we define the reveal matrix $Z_\sigma \in \mathbb{R}^{m \times m}$ as a diagonal matrix where $(Z_\sigma)_{ii} = 1$ if state $i \in \rho(\sigma)$ and $0$ otherwise. We define the transition matrix $T_\sigma \in \mathbb{R}^{m \times m}$ as a column-stochastic matrix where $(T_\sigma)_{ij} = \delta(j, \sigma)(i)$ represents the probability of transitioning to state $i$ from state $j$. The full belief update for input $\sigma_t$ is:
\begin{equation}\label{eq:beliefupdate}
    b_{t+1} = f(T_{\sigma_t} Z_{\sigma_t} b_t), \quad \text{where } f(x) = x / \|x\|_1
\end{equation}
This model captures two key aspects of code execution. First, partial observability: we receive reveals that prune possible states (e.g., through print statements) rather than observing the full state. Second, probabilistic transitions arise from operations with random choices or external inputs.

\textbf{Concrete example: unobservable branching as a PFSA-SR.}
To illustrate how code execution maps to a PFSA-SR, consider the unobservable branching scenario from Figure~\ref{fig:probabilistic-state-transitions}(A).
Suppose a program maintains a list \texttt{[a, b]} and executes \texttt{if random() < 0.5: a, b = b, a}. We model this as a PFSA-SR with $\mathcal{Q} = S_2 = \{(a,b),\, (b,a)\}$, two states corresponding to the two possible orderings. The conditional swap is a transition-only symbol $\sigma_{\mathrm{swap}}$ with $\rho(\sigma_{\mathrm{swap}}) = \mathcal{Q}$ (no information revealed) and $\delta(q, \sigma_{\mathrm{swap}})$ assigning probability $1/2$ to each state. A subsequent \texttt{print(a)} revealing that \texttt{a} equals its original value is a reveal-only symbol $\sigma_{\mathrm{rev}}$ with $\rho(\sigma_{\mathrm{rev}}) = \{(a,b)\}$ (only the identity ordering is consistent) and $\delta(q, \sigma_{\mathrm{rev}})(q) = 1$ (no state change). After observing $\sigma_{\mathrm{swap}}$ followed by $\sigma_{\mathrm{rev}}$, the belief collapses from $[1/2,\, 1/2]$ back to $[1,\, 0]$ via the reveal's support pruning and renormalization.

\textbf{Connections to existing models.} 
PFSA-SR subsumes and relates to several classical models. It reduces to a DFA when $\delta$ is deterministic and the reveals are not informative, i.e. $\rho (\sigma) = \mathcal{Q}$ for every $\sigma \in \Sigma$. If instead of hard observations we receive soft observations rather than hard constraints, the model is closely related to IO-HMMs~\citep{bengio1994input} or POMDP~\citep{aastrom1965optimal, kaelbling1998planning} observation models. If reveals are absent and $\delta$ is linear, we recover a Weighted Finite Automaton (WFA) viewpoint used in linear RNN theory. Compared to POMDPs and IO-HMMs, the essential difference here is the emphasis on hard, support-pruning reveals that eliminate some states rather than likelihood-weighted observations.

\subsection{Two Representations for Probabilistic Permutation Tracking}

To ground the PFSA-SR in a concrete but difficult problem, we analyze permutation tracking under uncertainty. We want to track the configurations of a list of $n$ distinct elements after a sequence of probabilistic permutations, each a convex combination of elements from group $S_n$. We consider two representations for the belief state, which differ in their computational complexity and stability under linear recurrence. 

\subsubsection{Joint Representation}

The \textit{joint representation} explicitly enumerates all possible automaton states. For permutation tracking over $S_n$, the state space contains $n!$ configurations (one for each permutation of the list). The belief $b_t \in \mathbb{R}^{n!}$ is a probability vector over these configurations. Note that this is much less efficient than the deterministic case, where we can track the state using a linear RNN with an $n-1$-dimensional state: each permutation in $S_n$ can be represented by a $n{-}1 \times n{-}1$ permutation matrix.

A key limitation of the joint representation in a linear recurrence is the role of the input injection term $\mB(\vx_t)$ (Eq.~\ref{eq:linrnn}). In the marginal representation (discussed below), a reveal directly replenishes decayed entries via $\mB$. In the joint representation, however, a reveal of the form ``position $i$ contains element $j$'' is consistent with $(n{-}1)!$ of the $n!$ configurations, and the \emph{subset} that survives depends on the current belief $b_t$, not just the input $\vx_t$. Since $\mB$ is a function of the input alone, it cannot adaptively identify which configurations to replenish. We return to this point in the stability analysis (\S\ref{sec:stability}), where we show it implies that the joint representation is always numerically unstable under repeated partial reveals. We provide a detailed worked example in~\Cref{app:example_linear_wfa}.

\subsubsection{Marginal Representation}

As a more tractable alternative, instead of tracking the \textit{joint probability} of all positions, we track the \textit{marginal probability} of each position independently. Consider a list of elements starting from $(1,2,\dots, n)$. The belief state is an $n \times n$ matrix $\mH_t$ where entry $(\mH_t)_{ij}$ represents the probability that position $i$ contains element $j$. The starting state is $\mH_0 = I$. To align with standard state-transition notation (left multiplication), rows ($i$) represent positions in memory and columns ($j$) represent elements/variables.

This matrix must satisfy two constraints derived from the definition of a permutation: (1) each position contains exactly one element, $\sum_{j} (\mH_t)_{ij} = 1$ for all $i$; and (2) each element exists at exactly one position, $\sum_{i} (\mH_t)_{ij} = 1$ for all $j$. Matrices satisfying these properties form the Birkhoff polytope of doubly stochastic matrices. Tracking the distribution over $S_n$ corresponds to moving a point within this polytope. By the Birkhoff--von Neumann theorem~\citep{birkhoff1946tres, von1953certain}, any doubly stochastic matrix can be expressed as a convex combination of permutation matrices.

The marginal representation admits a natural bi-linear recurrence:
\begin{equation*}
    \mH_t = \mA_l(\vx_t)\mH_{t-1}\mA_r(\vx_t) + \mB(\vx_t),\quad\hat{\vy}_t = \mathrm{dec}(\mH_t, \vx_t) \quad \text{where } t \in \{1, \dots, T\}
\end{equation*}
Using the identity (sometimes referred to as Roth's column lemma) $\mathrm{vec}(\mB\mX\mA^\top)=(\mA \otimes \mB)\mathrm{vec}(\mX)$~\citep{Roth1934OnDP,henderson1981vec}, we can see that this remains an affine update on the vectorized state: $\text{vec}(\mH_{t}) = (\mA_r(\vx_t)^\top \otimes \mA_l(\vx_t)) \text{vec}(\mH_{t-1}) + \text{vec}(\mB(\vx_t))$. We note that a similar two-sided recurrence structure is present in the open-source implementation of Kimi Delta Attention (KDA)~\citep{team2025kimi} \href{https://github.com/fla-org/flash-linear-attention/blob/75cc5aaa0ee121806dd5ca37ed733365f78aad1f/fla/ops/gated_delta_rule/fused_recurrent.py#L244C9-L244C10}{(code)}, although the recurrence was not explicitly described in the paper.

For representing the marginal distribution during probabilistic state-tracking, we parameterize the recurrence as $\mA_l (\vx_t) = \mP_{s} (\vx_t)\mD_l(\vx_t)$ and $\mA_r(\vx_t) = \mD_r(\vx_t)$, where $\mD_l(\vx_t), \mD_r(\vx_t)$ are diagonal matrices and $\mP_s(\vx_t)$ is a (possibly stochastic) permutation matrix.

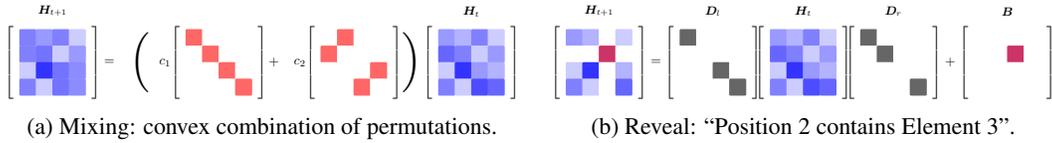
\begin{figure}[t]
\centering
\begin{subfigure}[t]{0.49\textwidth}
\centering
\begin{adjustbox}{width=\textwidth}
\begin{tikzpicture}[node distance=0.3cm]
    \tikzset{
        gridMatrix/.style={
            matrix of nodes,
            nodes={
                minimum size=0.45cm,
                anchor=center,
                inner sep=1pt,
                font=\scriptsize,
                rounded corners=1pt
            },
            column sep=-\pgflinewidth,
            row sep=-\pgflinewidth,
            nodes in empty cells,
            inner sep=3pt
        },
        op/.style={font=\normalsize, anchor=center}
    }
    \matrix[gridMatrix, left delimiter={[}, right delimiter={]}] (result) {
        |[fill=blue!50]| {} & |[fill=blue!40]| {} & |[fill=blue!50]| {} & |[fill=blue!20]| {} \\
        |[fill=blue!50]| {} & |[fill=blue!55]| {} & |[fill=blue!20]| {} & |[fill=blue!40]| {} \\
        |[fill=blue!20]| {} & |[fill=blue!80]| {} & |[fill=blue!55]| {} & |[fill=blue!45]| {} \\
        |[fill=blue!50]| {} & |[fill=blue!20]| {} & |[fill=blue!55]| {} & |[fill=blue!45]| {} \\
    };
    \node[above=3pt of result, font=\small] {$\mH_{t+1}$};
    \node[op, right=0.25cm of result] (eq) {$=$};
    \node[op, right=0.15cm of eq] (lparen) {\scalebox{2.5}{$\Big($}};    \node[op, right=0.002cm of lparen, font=] (c1) {$c_1$};
    \matrix[gridMatrix, right=0.15cm of c1, left delimiter={[}, right delimiter={]}] (p1) {
        |[fill=red!60]| {} & |[fill=white]| {} & |[fill=white]| {} & |[fill=white]| {} \\
        |[fill=white]| {} & |[fill=red!60]| {} & |[fill=white]| {} & |[fill=white]| {} \\
        |[fill=white]| {} & |[fill=white]| {} & |[fill=red!60]| {} & |[fill=white]| {} \\
        |[fill=white]| {} & |[fill=white]| {} & |[fill=white]| {} & |[fill=red!60]| {} \\
    };
    \node[op, right=0.2cm of p1] (plus) {$+$};
    \node[op, right=0.15cm of plus] (c2) {$c_2$};
    \matrix[gridMatrix, right=0.15cm of c2, left delimiter={[}, right delimiter={]}] (p2) {
        |[fill=white]| {} & |[fill=red!60]| {} & |[fill=white]| {} & |[fill=white]| {} \\
        |[fill=red!60]| {} & |[fill=white]| {} & |[fill=white]| {} & |[fill=white]| {} \\
        |[fill=white]| {} & |[fill=white]| {} & |[fill=white]| {} & |[fill=red!60]| {} \\
        |[fill=white]| {} & |[fill=white]| {} & |[fill=red!60]| {} & |[fill=white]| {} \\
    };
    \node[op, right=0.15cm of p2] (rparen) {\scalebox{2.5}{$\Big)$}};    \matrix[gridMatrix, right=0.2cm of rparen, left delimiter={[}, right delimiter={]}] (ht) {
        |[fill=blue!40]| {} & |[fill=blue!30]| {} & |[fill=blue!50]| {} & |[fill=blue!20]| {} \\
        |[fill=blue!60]| {} & |[fill=blue!50]| {} & |[fill=blue!20]| {} & |[fill=blue!40]| {} \\
        |[fill=blue!20]| {} & |[fill=blue!80]| {} & |[fill=blue!40]| {} & |[fill=blue!30]| {} \\
        |[fill=blue!50]| {} & |[fill=blue!20]| {} & |[fill=blue!70]| {} & |[fill=blue!60]| {} \\
    };
    \node[above=3pt of ht, font=\small] {$\mH_{t}$};
\end{tikzpicture}
\end{adjustbox}
\caption{Mixing: convex combination of permutations.}
\label{fig:mixing}
\end{subfigure}
\hfill
\begin{subfigure}[t]{0.48\textwidth}
\centering
\begin{adjustbox}{width=\textwidth}
\begin{tikzpicture}[node distance=0.3cm]
    \tikzset{
        gridMatrix/.style={
            matrix of nodes,
            nodes={
                minimum size=0.45cm,
                anchor=center,
                inner sep=1pt,
                font=\scriptsize,
                rounded corners=1pt
            },
            column sep=-\pgflinewidth,
            row sep=-\pgflinewidth,
            nodes in empty cells,
            inner sep=3pt
        },
        op/.style={font=\normalsize, anchor=center}
    }
    \matrix[gridMatrix, left delimiter={[}, right delimiter={]}] (result) {
        |[fill=blue!40]| {} & |[fill=blue!30]| {} & |[fill=white]| {} & |[fill=blue!20]| {} \\
        |[fill=white]| {} & |[fill=white]| {} & |[fill=purple!80]| {} & |[fill=white]| {} \\
        |[fill=blue!20]| {} & |[fill=blue!80]| {} & |[fill=white]| {} & |[fill=blue!30]| {} \\
        |[fill=blue!50]| {} & |[fill=blue!20]| {} & |[fill=white]| {} & |[fill=blue!60]| {} \\
    };
    \node[above=3pt of result, font=\small] {$\mH_{t+1}$};
    \node[op, right=0.25cm of result] (eq) {$=$};
    \matrix[gridMatrix, right=0.25cm of eq, left delimiter={[}, right delimiter={]}] (dl) {
        |[fill=black!60]| {} & |[fill=white]| {} & |[fill=white]| {} & |[fill=white]| {} \\
        |[fill=white]| {} & |[fill=white]| {} & |[fill=white]| {} & |[fill=white]| {} \\
        |[fill=white]| {} & |[fill=white]| {} & |[fill=black!60]| {} & |[fill=white]| {} \\
        |[fill=white]| {} & |[fill=white]| {} & |[fill=white]| {} & |[fill=black!60]| {} \\
    };
    \node[above=3pt of dl, font=\small] {$\mD_l$};
    \matrix[gridMatrix, right=0.4cm of dl, left delimiter={[}, right delimiter={]}] (ht) {
        |[fill=blue!40]| {} & |[fill=blue!30]| {} & |[fill=blue!50]| {} & |[fill=blue!20]| {} \\
        |[fill=blue!60]| {} & |[fill=blue!50]| {} & |[fill=blue!20]| {} & |[fill=blue!40]| {} \\
        |[fill=blue!20]| {} & |[fill=blue!80]| {} & |[fill=blue!40]| {} & |[fill=blue!30]| {} \\
        |[fill=blue!50]| {} & |[fill=blue!20]| {} & |[fill=blue!70]| {} & |[fill=blue!60]| {} \\
    };
    \node[above=3pt of ht, font=\small] {$\mH_{t}$};
    \matrix[gridMatrix, right=0.4cm of ht, left delimiter={[}, right delimiter={]}] (dr) {
        |[fill=black!60]| {} & |[fill=white]| {} & |[fill=white]| {} & |[fill=white]| {} \\
        |[fill=white]| {} & |[fill=black!60]| {} & |[fill=white]| {} & |[fill=white]| {} \\
        |[fill=white]| {} & |[fill=white]| {} & |[fill=white]| {} & |[fill=white]| {} \\
        |[fill=white]| {} & |[fill=white]| {} & |[fill=white]| {} & |[fill=black!60]| {} \\
    };
    \node[above=3pt of dr, font=\small] {$\mD_r$};
    \node[op, right=0.25cm of dr] (plus) {$+$};
    \matrix[gridMatrix, right=0.25cm of plus, left delimiter={[}, right delimiter={]}] (b) {
        |[fill=white]| {} & |[fill=white]| {} & |[fill=white]| {} & |[fill=white]| {} \\
        |[fill=white]| {} & |[fill=white]| {} & |[fill=purple!80]| {} & |[fill=white]| {} \\
        |[fill=white]| {} & |[fill=white]| {} & |[fill=white]| {} & |[fill=white]| {} \\
        |[fill=white]| {} & |[fill=white]| {} & |[fill=white]| {} & |[fill=white]| {} \\
    };
    \node[above=3pt of b, font=\small] {$\mB$};
\end{tikzpicture}
\end{adjustbox}
\caption{Reveal: ``Position 2 contains Element 3''.}
\label{fig:reveal}
\end{subfigure}
\caption{\textbf{Marginal state updates.} (a) Probabilistic mixing applies a convex combination of permutation matrices. (b) State reveals zero out conflicting row/column via diagonal masks and inject certainty via $\mB$.}
\label{fig:marginal_ops}
\end{figure}
\textbf{Probabilistic Transitions (Mixing).}
When the input $\vx_t$ encodes a shuffle (e.g., ``swap contents of Position $a$ and Position $b$''), we permute the rows of $\mH_t$. If the shuffle is probabilistic, we apply a linear mixture of permutation matrices:
\[ \mH_{t+1} = \mP_s(\vx_t) \mH_t,\qquad \mP_s(\vx_t) = \sum_{i=1}^{n!} c_i \mP_i,\qquad  \sum_{i=1}^{n!} c_i = 1\]
where $\mP_s$ acts on the Positions (rows) and $\mP_i$ are permutation matrices. The other components are set to $\mD_l(\vx_t) = \mD_r(\vx_t) = \mI$ and $\mB(\vx_t) = \mathbf{0}$. Note that $\mP_s(\vx_t)$ is doubly stochastic by definition, so if $\mH_{t}$ is doubly stochastic, $\mH_{t+1}$ will be as well.

\textbf{State Reveals.}
If $\vx_t$ encodes the constraint ``Position $i$ contains Element $j$'', we enforce: (1) entry $(i, j)$ is confirmed (set to $1$); (2) position $i$ cannot contain any other element (zero out row $i$ except column $j$); (3) element $j$ cannot be in any other position (zero out column $j$ except row $i$). This translates into:
\begin{equation*}
    \mP_s(\vx_t) = \mI,\quad \mD_l(\vx_t) = \mI - \ve_i\ve_i^\top,\quad \mD_r(\vx_t) = \mI - \ve_j\ve_j^\top,\quad \mB(\vx_t) = \ve_i \ve_j^\top
\end{equation*}
where $\ve_i$ is a standard basis vector of $\mathbb{R}^n$. The term $\mD_{l} \mH_t \mD_{r}$ preserves probability mass for configurations \textit{compatible} with the observation (outside the cross of row $i$ and column $j$), while $\mB(\vx_t)$ directly injects the certainty of the observed fact. After this update, $\mH_t$ may leave the Birkhoff polytope. To recover a valid marginal distribution, we apply the Sinkhorn-Knopp algorithm~\citep{sinkhorn1967concerning} in the decoder (to keep the main recursion linear), which iteratively normalizes rows and columns to project back onto the Birkhoff polytope. 

\subsection{Stability Analysis: When Does Each Representation Fail?}
\label{sec:stability}

We now characterize when linear RNNs can and cannot maintain stable belief tracking under each representation. The central issue is that the PFSA-SR belief update (Eq.~\ref{eq:beliefupdate}) requires nonlinear renormalization ($f(x) = x/\|x\|_1$) after each reveal. A linear RNN must defer this normalization to the decoder, propagating the \emph{unnormalized} state $h_t$ through time. We analyze the consequences for each representation.

\textbf{Joint representation: unstable under partial reveals.}
Consider the unnormalized linear recurrence that defers normalization:
\begin{equation}
\label{eq:unnorm_message}
h_{t+1} = T_{\sigma_t} Z_{\sigma_t} h_t,
\qquad 
b_t = h_t/\|h_t\|_1
\end{equation}
The ``survival probability'' at step $t$ is $s_t := \|Z_{\sigma_t} b_t\|_1 \in (0,1]$, representing the fraction of belief mass consistent with the reveal. The unnormalized state carries the cumulative product $\|h_{t}\|_1 = \prod_{k=0}^{t-1} s_k$ in its magnitude. Since $\mB(\vx_t)$ cannot replenish mass in the joint case (as discussed in \S4.2.1), whenever partial reveals repeatedly prune mass (i.e., $s_k<1$ for many steps), $\|h_t\|_1$ shrinks exponentially, and the state vanishes in finite precision. 
This is illustrated in Figure~\ref{fig:decay_viz}.

This is a known numerical
issue in HMM/Bayesian filtering: unnormalized forward messages $\alpha_t(i) = p(y_{1:t},x_t = i)$ encode
the observation-prefix likelihood in their scale and quickly underflow for long sequences, motivating
per-step scaling (“scaled forward-backward”) or log-domain implementations (e.g. log-sum-exp) for
numerical stability \citep{rabiner1989tutorial, murphy2002hidden}, both of which break the linearity of the recurrence.

\definecolor{ghostRed}{RGB}{255,220,220}
\definecolor{textRed}{RGB}{150,0,0}
\definecolor{revealGreen}{RGB}{220,255,220}
\definecolor{textGreen}{RGB}{0,100,0}

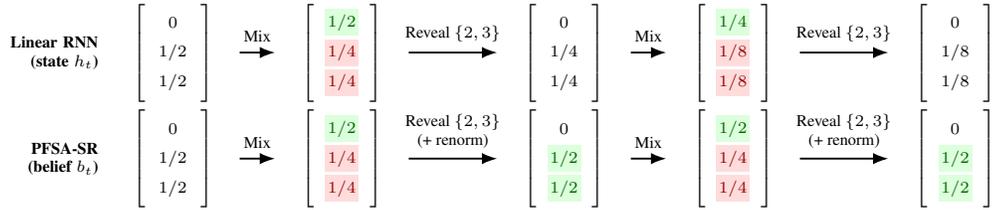
\begin{figure}[t]
\centering
\noindent
\resizebox{\linewidth}{!}{%
\begin{tikzpicture}[
    font=\scriptsize,
    >=Latex,
    vec/.style={
        matrix of math nodes,
        nodes={anchor=center, minimum width=5mm, minimum height=4mm, inner sep=0pt},
        left delimiter={[}, right delimiter={]},
        row sep=1.5pt,
        ampersand replacement=\&
    },
    op arrow/.style={->, thick, shorten >=18pt, shorten <=18pt},
    lbl/.style={font=\scriptsize, above, yshift=3pt, fill=white, inner sep=1.2pt, align=center},
    rowlabel/.style={font=\scriptsize\bfseries, align=right, text width=22mm},
    notelabel/.style={font=\scriptsize, align=center, text=gray}
]

\node[rowlabel, anchor=east] (Llab) at (0,0) {Linear RNN\\(state $h_t$)};

\matrix (m0) [vec] at (10mm,0) {
    0\\ 1/2\\ 1/2\\
};

\matrix (m1mix) [vec, right=18mm of m0] {
    |[fill=revealGreen, text=textGreen]| 1/2\\
    |[fill=ghostRed, text=textRed]| 1/4\\
    |[fill=ghostRed, text=textRed]| 1/4\\
};

\matrix (m1rev) [vec, right=26mm of m1mix] {
    0\\
    1/4\\
    1/4\\
};

\matrix (m2mix) [vec, right=18mm of m1rev] {
    |[fill=revealGreen, text=textGreen]| 1/4\\
    |[fill=ghostRed, text=textRed]| 1/8\\
    |[fill=ghostRed, text=textRed]| 1/8\\
};

\matrix (m2rev) [vec, right=26mm of m2mix] {
    0\\
    1/8\\
    1/8\\
};

\draw[op arrow] (m0.east)    -- node[lbl] {Mix} (m1mix.west);
\draw[op arrow] (m1mix.east) -- node[lbl] {Reveal $\{2,3\}$} (m1rev.west);
\draw[op arrow] (m1rev.east) -- node[lbl] {Mix} (m2mix.west);
\draw[op arrow] (m2mix.east) -- node[lbl] {Reveal $\{2,3\}$} (m2rev.west);

\node[rowlabel, anchor=east] (Nlab) at (0,-16mm) {PFSA-SR\\(belief $b_t$)};

\matrix (b0) [vec] at (10mm,-16mm) {
    0\\ 1/2\\ 1/2\\
};

\matrix (b1mix) [vec, right=18mm of b0] {
    |[fill=revealGreen, text=textGreen]| 1/2\\
    |[fill=ghostRed, text=textRed]| 1/4\\
    |[fill=ghostRed, text=textRed]| 1/4\\
};

\matrix (b1rev) [vec, right=26mm of b1mix] {
    0\\
    |[fill=revealGreen, text=textGreen]| 1/2\\
    |[fill=revealGreen, text=textGreen]| 1/2\\
};

\matrix (b2mix) [vec, right=18mm of b1rev] {
    |[fill=revealGreen, text=textGreen]| 1/2\\
    |[fill=ghostRed, text=textRed]| 1/4\\
    |[fill=ghostRed, text=textRed]| 1/4\\
};

\matrix (b2rev) [vec, right=26mm of b2mix] {
    0\\
    |[fill=revealGreen, text=textGreen]| 1/2\\
    |[fill=revealGreen, text=textGreen]| 1/2\\
};

\draw[op arrow] (b0.east)    -- node[lbl] {Mix} (b1mix.west);
\draw[op arrow] (b1mix.east) -- node[lbl] {Reveal $\{2,3\}$\\(+ renorm)} (b1rev.west);
\draw[op arrow] (b1rev.east) -- node[lbl] {Mix} (b2mix.west);
\draw[op arrow] (b2mix.east) -- node[lbl] {Reveal $\{2,3\}$\\(+ renorm)} (b2rev.west);

\end{tikzpicture}%
}
\caption{\textbf{Joint representation instability.} A mixing step moves half the mass into state 1 (absorbing): from states 2 and 3 we transition to state 1 with probability $1/2$ and otherwise stay put. The reveal keeps $\{2,3\}$. A linear recurrence storing the unnormalized state $h_t$ loses mass exponentially fast ($1/2\!\to\!1/4\!\to\!1/8\!\to\!\cdots$), so all entries vanish in finite precision. PFSA-SR normalizes after each reveal, keeping two entries bounded away from zero ($1/2,1/2$).}
\label{fig:decay_viz}
\end{figure}

    \definecolor{ghostRed}{RGB}{255,220,220}
    \definecolor{textRed}{RGB}{150,0,0}
    \definecolor{revealGreen}{RGB}{220,255,220}
    \definecolor{textGreen}{RGB}{0,100,0}

\begin{figure}
    \centering
\adjustbox{width=1.0\linewidth, valign=t}{
\begin{tikzpicture}[
    node distance=1.5cm,
    font=\scriptsize,
    >=latex,
    mat/.style={
        matrix of math nodes,
        nodes={anchor=center, minimum width=0.2cm, minimum height=0.2cm, inner sep=0pt},
        left delimiter=(, right delimiter=),
        column sep=1.5pt, row sep=1.5pt,
        ampersand replacement=\&
    },
    op arrow/.style={->, thick, shorten >=9pt, shorten <=9pt},
    note/.style={font=\tiny\itshape, align=center, color=gray}
]

    \matrix (h0) [mat] {
        1 \& 0 \& 0 \\
        0 \& 1 \& 0 \\
        0 \& 0 \& 1 \\
    };
    
    \matrix (h1) [mat, right=of h0] {
        1 \& 0 \& 0 \\
        0 \& |[fill=ghostRed, text=textRed]| 0.5 \& |[fill=ghostRed, text=textRed]| 0.5 \\
        0 \& |[fill=ghostRed, text=textRed]| 0.5 \& |[fill=ghostRed, text=textRed]| 0.5 \\
    };
    
    \matrix (h2) [mat, right=of h1] {
        1 \& 0 \& 0 \\
        0 \& |[fill=revealGreen, text=textGreen]| 1 \& 0 \\
        0 \& 0 \& |[fill=ghostRed, text=textRed]| 0.5 \\
    };
    
    \matrix (h3) [mat, right=of h2] {
        1 \& 0 \& 0 \\
        0 \& |[fill=ghostRed, text=textRed]| 0.5 \& |[fill=ghostRed, text=textRed]| 0.25 \\
        0 \& |[fill=ghostRed, text=textRed]| 0.5 \& |[fill=ghostRed, text=textRed]| 0.25 \\
    };
    
    \matrix (h4) [mat, right=of h3] {
        1 \& 0 \& 0 \\
        0 \& |[fill=revealGreen, text=textGreen]| 1 \& 0 \\
        0 \& 0 \& |[fill=ghostRed, text=textRed]| 0.25 \\
    };
    
    \matrix (h5) [mat, right=of h4] {
        1 \& 0 \& 0 \\
        0 \& |[fill=ghostRed, text=textRed]| 0.5 \& |[fill=ghostRed, text=textRed]| 0.125 \\
        0 \& |[fill=ghostRed, text=textRed]| 0.5 \& |[fill=ghostRed, text=textRed]| 0.125 \\
    };

    \draw[op arrow] (h0) -- node[above] {$ 2 \stackrel{?}{\leftrightarrow} 3 $} (h1);
    \draw[op arrow] (h1) -- node[above] {$R(2,2)$} (h2);
    \draw[op arrow] (h2) -- node[above] {$ 2 \stackrel{?}{\leftrightarrow} 3 $} (h3);
    \draw[op arrow] (h3) -- node[above] {$R(2,2)$} (h4);
    \draw[op arrow] (h4) -- node[above] {$ 2 \stackrel{?}{\leftrightarrow} 3 $} (h5);

    \foreach \i in {0,...,5} {
        \node[below=0.1cm of h\i] {$\mH_\i$};
    }
\end{tikzpicture}}
\vspace{-3mm}
    \caption{Visualizing the exponential decay of unnormalized probability mass in a Linear RNN for the \textbf{marginal state}. We start with the identity state ($\mH_0$). A probabilistic mixing of elements 2 and 3 ($\mH_1$) followed by a deterministic reveal of element 2 at position 2 ($\mH_2$) correctly resets the revealed entry to 1 but leaves the unrevealed entry at 0.5. Repeating this cycle ($\mH_3, \mH_4, \mH_5$) causes the unrevealed mass to decay exponentially ($0.5 \to 0.25 \to 0.125$), eventually vanishing in finite precision.}
    \label{fig:decay_viz_2}
\end{figure}

\paragraph{Marginal representation: stable with sufficient reveals, unstable under adversarial sequences.}
In contrast, the marginal representation can leverage $\mB(\vx_t)$ to replenish decaying entries. When a reveal ``position $i$ contains element $j$'' occurs, $\mB(\vx_t) = \ve_i \ve_j^\top$ directly injects a 1 at entry $(i,j)$, regardless of its prior value. This acts as a ``reset'' that prevents accumulated decay.

Stable case: If each position/element pair is revealed periodically, every entry in the marginal matrix $\mH_t$ receives periodic replenishment. The decay between reveals is bounded, and finite precision suffices.

Adversarial case: An adversarial reveal sequence can cause specific entries to decay indefinitely. Consider $n=3$ elements starting at identity $\mH_0=I$. The adversary repeatedly: (1) applies a probabilistic swap mixing elements 2 and 3, then (2) reveals ``position 2 contains element 2''. The reveal replenishes entry $(2,2)$ to 1, but element 3's mass (at positions 2 and 3) is never directly revealed, it can only be inferred by exclusion. Each cycle halves the unrevealed mass ($1 \to 0.5 \to 0.25 \to \cdots$) until it vanishes in finite precision. See Figure~\ref{fig:decay_viz_2} for an illustration.

\textbf{Summary.} Under the representations analyzed here, the joint representation is numerically unstable for linear RNNs because reveals cannot replenish mass, and the marginal representation can be stable if reveals are distributed across all variables but admits adversarial sequences that force exponential decay. These constructions provide strong evidence that linear RNNs face an inherent difficulty with probabilistic state-tracking: without either nonlinear renormalization in the recurrence or constraints on the reveal distribution, stable belief maintenance appears infeasible.

One might note that finite precision makes the set of representable beliefs finite, so in principle the dynamics can be simulated by a DFA (and hence linearized via one-hot encoding). However, this is impractical: in the joint case, even a binary discretization of the $n!$-dimensional belief vector requires $2^{n!}$ DFA states; in the marginal case, the belief lives in the Birkhoff polytope of dimension $(n{-}1)^2$, so discretizing each coordinate into $k$ bins yields $k^{(n-1)^2}$ states (e.g., $10^{81}$ for $n{=}10$, $k{=}10$).

\textbf{Linear RNNs suffice for deterministic automata and full state reveals.}
Finally, we note two regimes where Linear RNNs suffice. First, if the system is a \emph{Deterministic} Finite Automaton (DFA), the belief state remains a one-hot vector corresponding to the true state. Since no uncertainty is introduced, the normalization constant is always 1, and the update is linear. Second, if the system receives \emph{Full State Reveals} (observing the exact current state), the belief is effectively reset to a one-hot vector. This "hard reset" can be implemented by the linear update $h_{t+1} = 0 \cdot h_t + \mathbf{c}$, which zeroes out the history and injects unit mass at the revealed state. Consequently, in a probabilistic setting with partial and full reveals, if full state reveals occur frequently, they prevent the exponential decay of the state norm caused by partial reveals stabilizing the recurrence.

\section{Conclusion}
We study state-tracking in neural sequence models through the lens of code execution, bridging the gap between abstract automata benchmarks and the next-token prediction paradigm used to train language models.

\textbf{Summary of contributions.}
First, we introduced Python REPL traces as a testbed for state-tracking under next-token prediction, showing that linear RNNs with extended eigenvalue spectra (DeltaNet$[-1,1]$) can learn and generalize reliably with sparse state supervision, while Transformers fail even with dense reveals.
Second, we provided evidence for a barrier facing linear RNNs in realistic code settings: when transitions are probabilistic or partially observable, exact belief tracking requires nonlinear renormalization. We formalized this through the PFSA-SR framework and exhibited adversarial reveal sequences that cause exponential decay in state norms under the natural joint and marginal representations, suggesting that stable belief maintenance is infeasible for linear RNNs under finite precision.

\textbf{Limitations.}
Our experiments focus on synthetic permutation groups ($S_n$) rather than real-world code. While REPL traces are more realistic than prior sequence-to-sequence setups, they still simplify away parsing, control flow, and memory management.

\textbf{Future work.}
Our findings suggest that probabilistic state-tracking is a promising benchmark for evaluating nonlinear RNNs~\citep{hochreiter1997long, beck2024xlstm} and recent parallelization efforts~\citep{limparallelizing, gonzalez2024towards, danieli2025pararnn}. Extending this work to real execution traces, e.g. from CRUXEval or system call logs, and investigating hybrid architectures that interleave linear recurrence with periodic nonlinear normalization similar to TTT~\citep{sunlearning} and Titans~\citep{behrouz2025titans} is left as future work. Concurrent work demonstrates that Kalman-style filtering can be parallelized for language modeling~\citep{shaj2026kalman}.

\section*{Acknowledgments}
We would like to thank Jan Tönshoff, Heiner Kremer, Fabian Falck, Alicia Curth, Teodora Pandeva, Hari Govind V K, and Andrey Rybalchenko for insightful discussions throughout the project.

\section*{Author Contributions}
J.S. proposed to use code execution traces in the form of REPL traces and iterated on the format with R.G., K.K. and B.R. Identified failure cases under partial observability in code based on execution traces, conceived the possibility for failure of probabilistic state-tracking in linear RNNs for the joint representation in this setting. Developed the software implementation, ran the primary experiments, and led the manuscript writing. R.G. iterated on the REPL traces format, formalized the probabilistic finite-state example by J.S, proposed marginal state-representations and made connection to prior work on recurrences, and contributed heavily to the manuscript writing. K.P. did preliminary experiments on comparing linear and non-linear RNNs comparing their practical probabilistic state tracking capabilities. K.K., H.B., and B.R. provided project supervision, iterated on the format of the REPL traces, guided the overarching research direction, and contributed to manuscript revisions. B.R. provided additional conceptual framing based on parallel work in code world models.

\bibliography{iclr2025_conference}
\bibliographystyle{iclr2025_conference}

\newpage
\appendix

\section{From Permutation Groups to Variable Tracking in Code}

\subsection{Experimental Details}\label{app:experimental_details}
\textbf{Model \& Training.} We train a DeltaNet and Transformer ($\approx$265M parameters) using the implementation from flash-linear-attention~\citep{yang2024fla}. The architecture consists of 18 layers, hidden dimension $d=512$, 8 heads ($d_{head}=128$), MLP expansion factor 4, and SwiGLU activations. Optimization is performed using AdamW~\citep{loshchilovdecoupled} with a peak learning rate of $5\times 10^{-4}$, a cosine decay schedule~\citep{loshchilov2017sgdr} (minimum LR ratio 0.2), and 5\% warmup. We use a per-device batch size of 3 with 12 gradient accumulation steps in BF16 mixed precision. We trained on single Nvidia A100s for each model.

\textbf{Curriculum Learning.} To ensure stability, we use a four-stage curriculum of 15,000 samples each. We use full permutations on 5 variables ($S_5$), progressively increasing the trace length ($L$) and reveal spacing ($S$):
$$(L, S) \in \{(8, 1), (16, 2), (32, 4), (64, 8)\}$$

\newpage
\subsection{Interpretability}\label{app:interpretability}

\begin{figure}[h]
    \centering
    \adjustbox{width=0.85\textwidth}{
    \includegraphics[width=0.313\linewidth, trim=0 0 0 0, clip=true]{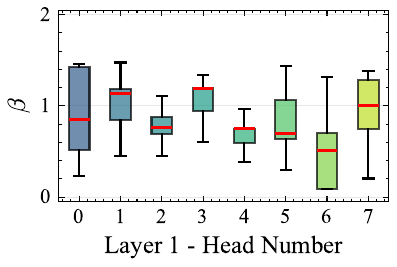}
    \hspace{1mm}
    \includegraphics[width=0.28\linewidth, trim=20 0 0 0, clip=true]{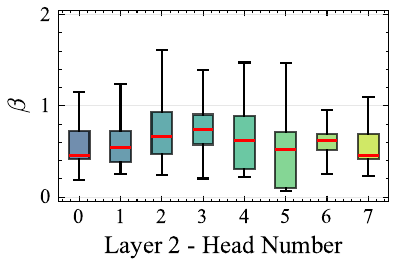}
    \hspace{1mm}
    \includegraphics[width=0.28\linewidth, trim=20 0 0 0, clip=true]{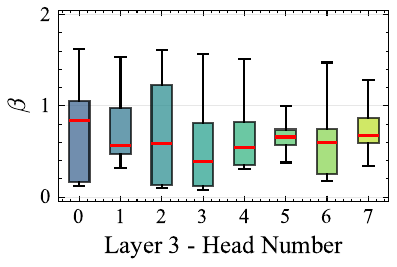}}
    \adjustbox{width=0.85\textwidth}{
    \includegraphics[width=0.313\linewidth, trim=0 0 0 0, clip=true]{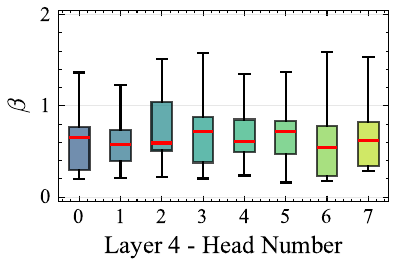}
    \hspace{1mm}
    \includegraphics[width=0.28\linewidth, trim=20 0 0 0, clip=true]{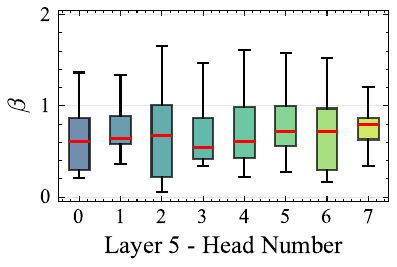}
    \hspace{1mm}
    \includegraphics[width=0.28\linewidth, trim=20 0 0 0, clip=true]{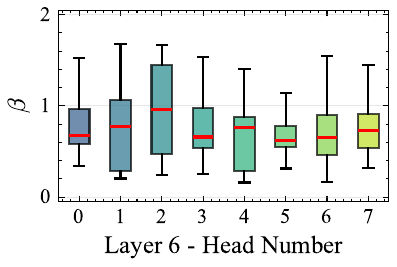}}
    \adjustbox{width=0.85\textwidth}{
    \includegraphics[width=0.313\linewidth, trim=0 0 0 0, clip=true]{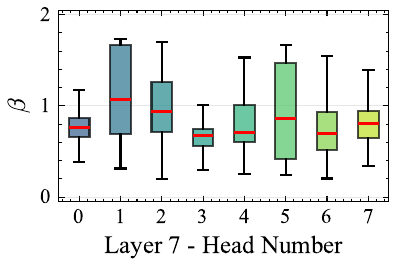}
    \hspace{1mm}
    \includegraphics[width=0.28\linewidth, trim=20 0 0 0, clip=true]{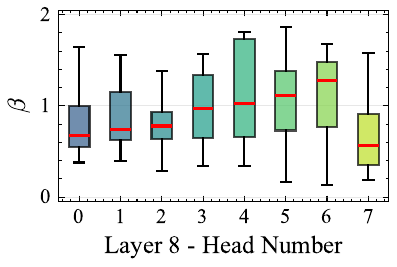}
    \hspace{1mm}
    \includegraphics[width=0.28\linewidth, trim=20 0 0 0, clip=true]{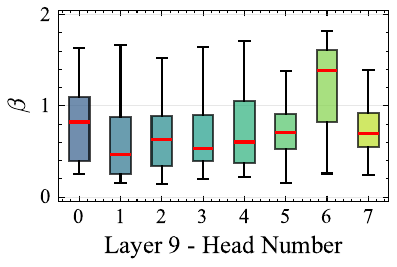}}
    \adjustbox{width=0.85\textwidth}{
    \includegraphics[width=0.313\linewidth, trim=0 0 0 0, clip=true]{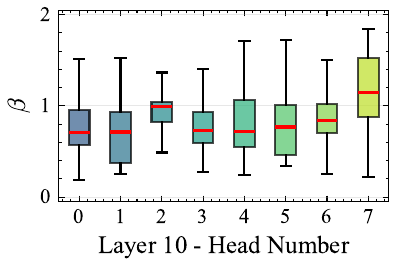}
    \hspace{1mm}
    \includegraphics[width=0.28\linewidth, trim=20 0 0 0, clip=true]{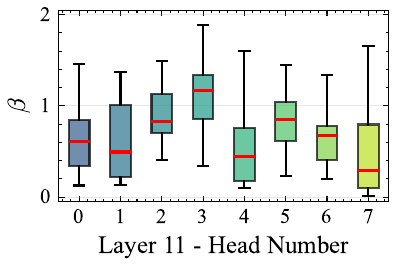}
    \hspace{1mm}
    \includegraphics[width=0.28\linewidth, trim=20 0 0 0, clip=true]{figures/motivation/interpretability/layer_model_layers_11_attn_b_proj_heads_boxplot_sample_0.pdf}}
    \adjustbox{width=0.85\textwidth}{
    \includegraphics[width=0.313\linewidth, trim=0 0 0 0, clip=true]{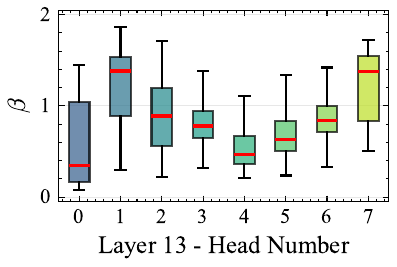}
    \hspace{1mm}
    \includegraphics[width=0.28\linewidth, trim=20 0 0 0, clip=true]{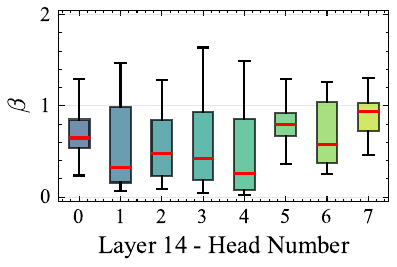}
    \hspace{1mm}
    \includegraphics[width=0.28\linewidth, trim=20 0 0 0, clip=true]{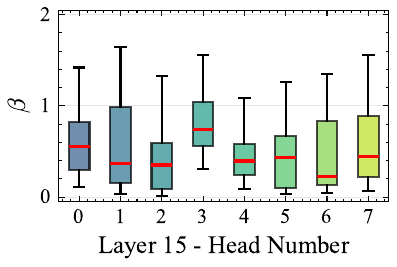}}
    \adjustbox{width=0.85\textwidth}{
    \includegraphics[width=0.313\linewidth, trim=0 0 0 0, clip=true]{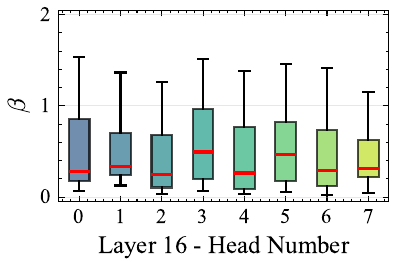}
    \hspace{1mm}
    \includegraphics[width=0.28\linewidth, trim=20 0 0 0, clip=true]{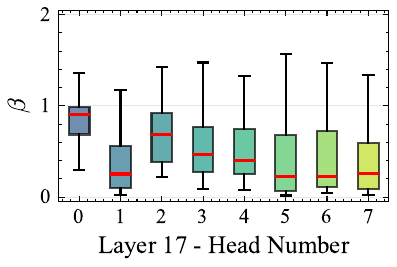}
    \hspace{1mm}
    \includegraphics[width=0.28\linewidth, trim=20 0 0 0, clip=true]{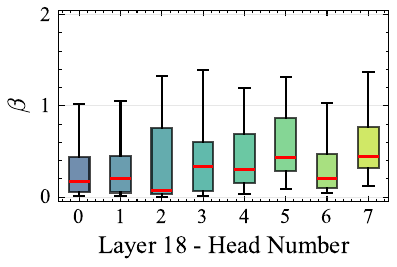}}
    \adjustbox{width=0.85\textwidth}{
    \includegraphics[width=0.313\linewidth, trim=0 0 0 0, clip=true]{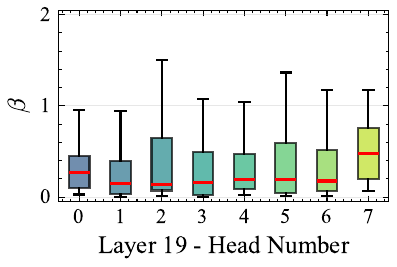}
    \hspace{1mm}
    \includegraphics[width=0.28\linewidth, trim=20 0 0 0, clip=true]{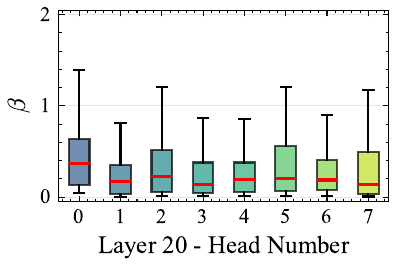}
    \hspace{1mm}
    \includegraphics[width=0.28\linewidth, trim=20 0 0 0, clip=true]{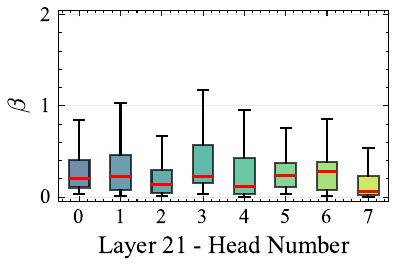}}
    \caption{$\beta$ distribution per layer per head recorded during a forward pass of a REPL trace. Layer 12, Head 3 stands out as the state-tracking head.}
    \label{fig:beta_appendix_motivation}
\end{figure}

\begin{figure}[h]
    \centering
    \adjustbox{width=0.85\textwidth}{
    \includegraphics[width=0.313\linewidth, trim=0 0 0 0, clip=true]{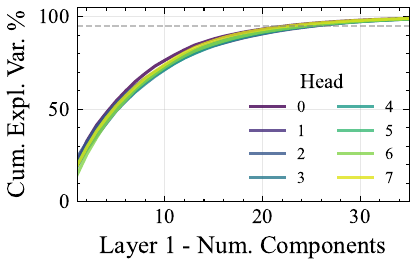}
    \hspace{1mm}
    \includegraphics[width=0.28\linewidth, trim=20 0 0 0, clip=true]{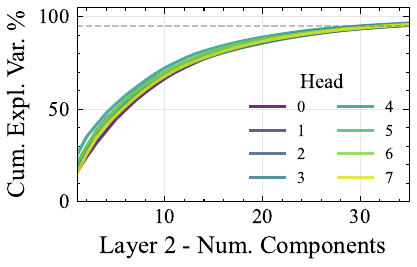}
    \hspace{1mm}
    \includegraphics[width=0.28\linewidth, trim=20 0 0 0, clip=true]{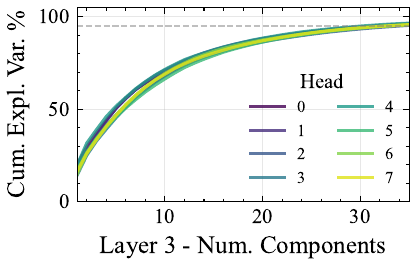}}   
    \adjustbox{width=0.85\textwidth}{
    \includegraphics[width=0.313\linewidth, trim=0 0 0 0, clip=true]{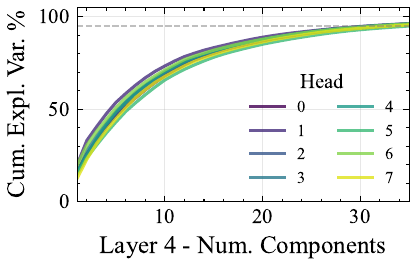}
    \hspace{1mm}
    \includegraphics[width=0.28\linewidth, trim=20 0 0 0, clip=true]{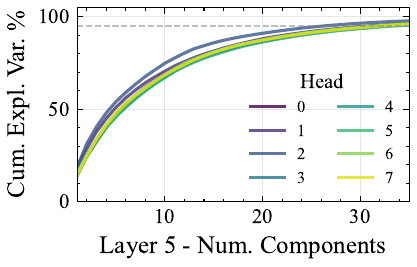}
    \hspace{1mm}
    \includegraphics[width=0.28\linewidth, trim=20 0 0 0, clip=true]{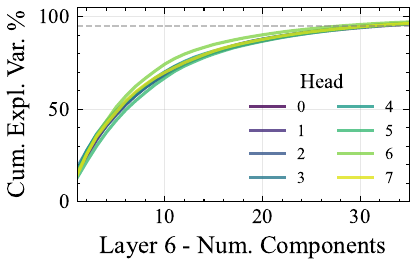}} 
    \adjustbox{width=0.85\textwidth}{
    \includegraphics[width=0.313\linewidth, trim=0 0 0 0, clip=true]{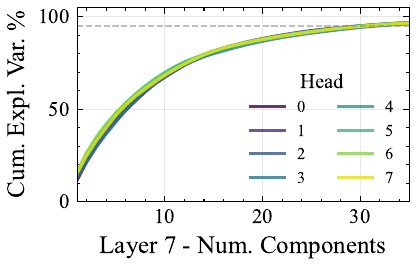}
    \hspace{1mm}
    \includegraphics[width=0.28\linewidth, trim=20 0 0 0, clip=true]{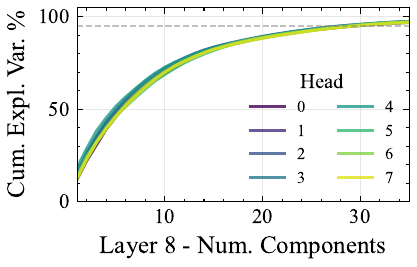}
    \hspace{1mm}
    \includegraphics[width=0.28\linewidth, trim=20 0 0 0, clip=true]{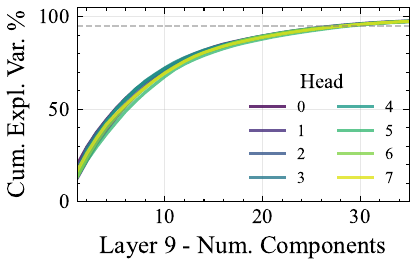}} 
    \adjustbox{width=0.85\textwidth}{
    \includegraphics[width=0.313\linewidth, trim=0 0 0 0, clip=true]{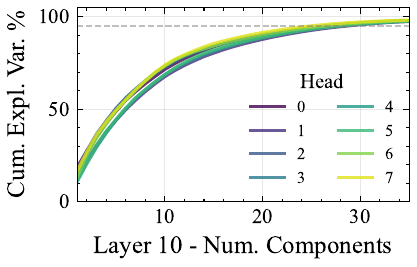}
    \hspace{1mm}
    \includegraphics[width=0.28\linewidth, trim=20 0 0 0, clip=true]{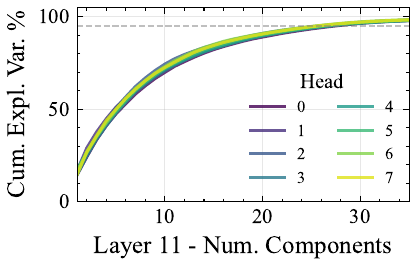}
    \hspace{1mm}
    \includegraphics[width=0.28\linewidth, trim=20 0 0 0, clip=true]{figures/motivation/interpretability/layer_11_key_pca_sample_0.pdf}} 
    \adjustbox{width=0.85\textwidth}{
    \includegraphics[width=0.313\linewidth, trim=0 0 0 0, clip=true]{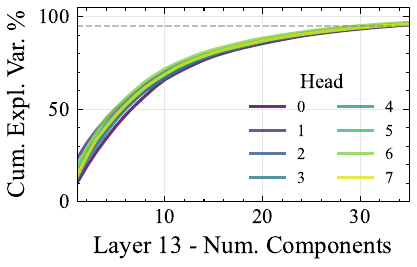}
    \hspace{1mm}
    \includegraphics[width=0.28\linewidth, trim=20 0 0 0, clip=true]{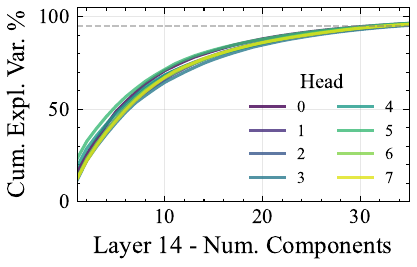}
    \hspace{1mm}
    \includegraphics[width=0.28\linewidth, trim=20 0 0 0, clip=true]{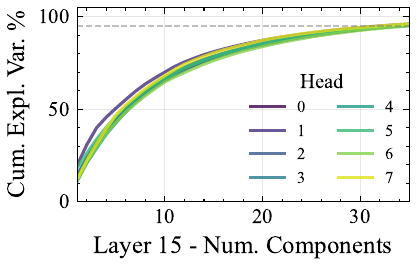}} 
    \adjustbox{width=0.85\textwidth}{
    \includegraphics[width=0.313\linewidth, trim=0 0 0 0, clip=true]{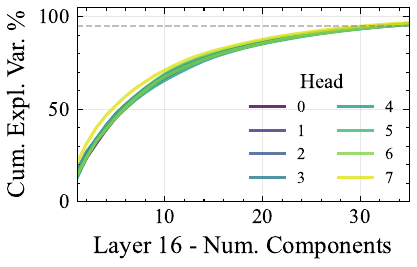}
    \hspace{1mm}
    \includegraphics[width=0.28\linewidth, trim=20 0 0 0, clip=true]{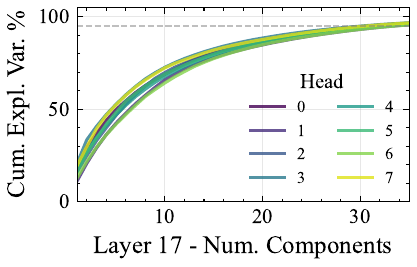}
    \hspace{1mm}
    \includegraphics[width=0.28\linewidth, trim=20 0 0 0, clip=true]{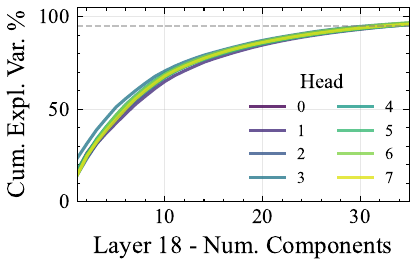}} 
    \adjustbox{width=0.85\textwidth}{
    \includegraphics[width=0.313\linewidth, trim=0 0 0 0, clip=true]{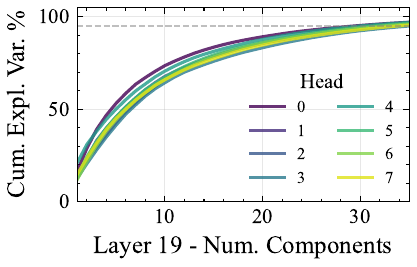}
    \hspace{1mm}
    \includegraphics[width=0.28\linewidth, trim=20 0 0 0, clip=true]{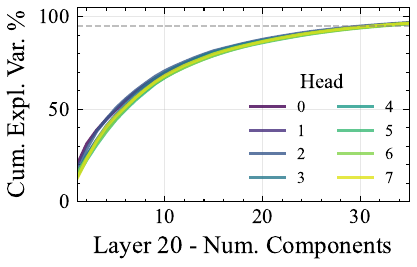}
    \hspace{1mm}
    \includegraphics[width=0.28\linewidth, trim=20 0 0 0, clip=true]{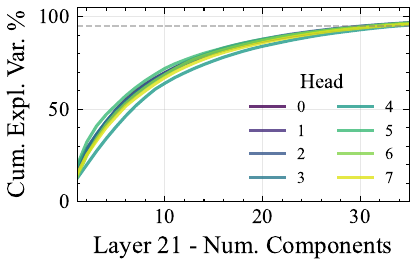}} 
    \caption{Cumulative explained variance of the keys per layer per head recorded during a forward pass of a REPL trace. Layer 12, Head 3 stands out as the state-tracking head.}
    \label{fig:placeholder}
\end{figure}
\newpage

\section{Probabilistic State-Tracking}
\subsection{Example: Linear RNN implementing probabilistic finite-state automaton tracking the join}\label{app:example_linear_wfa}

This section details the explicit arithmetic of a stochastic update on the permutation group $S_3$ and discusses the theoretical implications of norm decay in Linear RNNs.

\textbf{Scenario Setup and Initialization ($t=0$).}
We model the system state $\mH_t \in \R^6$ over the permutation group $S_3$. The basis vectors (universes) correspond to the six possible permutations of $[1,2,3]$: Identity $[1,2,3]$ (Index 1), Swap 1-2 $[2,1,3]$ (Index 2), Swap 1-3 $[3,2,1]$ (Index 3), Swap 2-3 $[1,3,2]$ (Index 4), Cycle Left $[2,3,1]$ (Index 5), and Cycle Right $[3,1,2]$ (Index 6). We begin at Step 0 with perfect certainty at the Identity configuration:
$$
\mH_0 = \begin{bmatrix} 1 & 0 & 0 & 0 & 0 & 0 \end{bmatrix}^T \quad (\text{Universe 1: } [1, 2, 3]).
$$

\textbf{Step 1: The Stochastic Action ($t=1$).}
The system receives the input "Try to Swap 1-2" with a noise profile defined as a 50\% chance of the intended Swap 1-2 and a 50\% chance of an accidental Swap 1-3. To represent this ambiguity, we construct a "fuzzy" transition matrix $\mA_{fuzzy}$ by averaging the permutation matrices of the two outcomes: $\mA_{fuzzy} = 0.5 \cdot \mA_{\text{swap12}} + 0.5 \cdot \mA_{\text{swap13}}$. Specifically, $\mA_{\text{swap12}}$ maps Identity to Index 2, while $\mA_{\text{swap13}}$ maps Identity to Index 3. The resulting update is:
$$
\mH_1 = \mA_{fuzzy} \mH_0 = 
\begin{bmatrix}
0   & 0.5 & 0.5 & 0   & 0   & 0   \\
\mathbf{0.5} & 0   & 0   & 0   & 0.5 & 0   \\
\mathbf{0.5} & 0   & 0   & 0   & 0   & 0.5 \\
0   & 0   & 0   & 0   & 0.5 & 0.5 \\
0   & 0.5 & 0   & 0.5 & 0   & 0   \\
0   & 0   & 0.5 & 0.5 & 0   & 0
\end{bmatrix}
\begin{bmatrix} 1 \\ 0 \\ 0 \\ 0 \\ 0 \\ 0 \end{bmatrix}
= \begin{bmatrix} 0 \\ \mathbf{0.5} \\ \mathbf{0.5} \\ 0 \\ 0 \\ 0 \end{bmatrix}.
$$
The state is now diffused; probability mass is split between Universe 2 ($[2,1,3]$) and Universe 3 ($[3,2,1]$).

\textbf{Step 2: The Observation ($t=2$).}
We subsequently receive the observation "Position 1 contains Object 3". To process this, we construct a diagonal observation matrix $\mA_{obs}$ that acts as a filter. We check every basis vector: Index 1 ($[1,2,3]$) is false; Index 2 ($[2,1,3]$) is false; Index 3 ($[3,2,1]$) is true (keep); Index 6 ($[3,1,2]$) is true (keep). All others are zeroed out. Applying this filter to the smeared state $\mH_1$:
$$
\mH_2 = \mA_{obs} \mH_1 = 
\begin{bmatrix}
0 & 0 & 0 & 0 & 0 & 0 \\
0 & 0 & 0 & 0 & 0 & 0 \\
0 & 0 & \mathbf{1} & 0 & 0 & 0 \\
0 & 0 & 0 & 0 & 0 & 0 \\
0 & 0 & 0 & 0 & 0 & 0 \\
0 & 0 & 0 & 0 & 0 & \mathbf{1}
\end{bmatrix}
\begin{bmatrix} 0 \\ 0.5 \\ 0.5 \\ 0 \\ 0 \\ 0 \end{bmatrix}
= \begin{bmatrix} 0 \\ 0 \\ \mathbf{0.5} \\ 0 \\ 0 \\ 0 \end{bmatrix}.
$$
The model has correctly identified that we are in Universe 3 ($[3,2,1]$). The ambiguity created in Step 1 was resolved because Universe 2 is inconsistent with the observation. The final vector magnitude is $0.5$, representing the joint probability of the path: $P(\text{Path}) = P(\text{Slip}) \times P(\text{Consistent}) = 0.5 \times 1.0 = 0.5$.

\textbf{The Mechanics of Information Decay.}
A critical limitation of Linear RNNs when tracking probabilistic states is the phenomenon of \emph{norm decay}. Unlike non-linear models (e.g., Transformers with Softmax) which re-normalize their internal state, a Linear RNN performs purely multiplicative updates ($\mH_t = \mA \mH_{t-1}$). In a stochastic setting, transition matrices often have eigenvalues $|\lambda| < 1$ due to diffusion or filtering. Consequently, $\|\mH_t\| \approx \lambda^t \|\mH_0\|$. For example, in a "Noisy Identity" transition where the system retains state with $p=0.9$, the norm decays to $\approx 0.00002$ after 100 steps. While floating-point standards allow for small numbers, the signal eventually vanishes relative to numerical noise.

\textbf{The Role of B: The Gated Reset.}
The matrix $\mB$, which is typically zero during standard tracking, serves as a solution to decay via a \emph{Gated Reset Mechanism}. Let $\vx_t$ be a binary indicator where $\vx_t=1$ indicates a "Reset". We parameterize the weights as $\mA(\vx_t) = (1 - \vx_t) \cdot \mA_{\text{step}}$ and $\mB(\vx_t) = \vx_t \cdot \mH_{\text{prior}}$.
When a reset is triggered ($\vx_t=1$), the history is annihilated ($\mA(\vx_t) = \mathbf{0}$) and the bias term injects the prior ($\mB(\vx_t) = \mH_{\text{prior}}$). The update becomes:
$$ \mH_t = \mathbf{0} \cdot \mH_{t-1} + \mH_{\text{prior}} = [1/6, \dots, 1/6]^T. $$
This "re-inflates" the state vector to full magnitude, readying the system to track a new sequence.

\end{document}